\documentclass[11pt,a4paper]{article}
\usepackage[table]{xcolor}
\usepackage[hyperref]{emnlp2020}
\usepackage{times}
\usepackage{latexsym}

 \usepackage{amssymb}
 \usepackage{amsmath}
 \usepackage{subcaption}
\usepackage[ruled,vlined,linesnumbered]{algorithm2e}
\SetKwInput{KwInput}{Input}                
\SetKwInput{KwOutput}{Output}              

\usepackage{microtype}
\usepackage{graphicx}

\usepackage{collcell}
\usepackage{pgf}
\usepackage{multirow}
\usepackage{multicol} 
\usepackage{booktabs}

\usepackage{colortbl}
\usepackage{tabto}

\aclfinalcopy 

\definecolor{myblue}{rgb}{0.9, 0.1, 0.94}
\definecolor{mygreen}{rgb}{0.64, 0.76, 0.68}
\definecolor{myyellow}{rgb}{0.88, 0.54, 0.35}
\definecolor{mygreen}{rgb}{0.68, 0.85, 0.9}

\newenvironment{itemize*}%
  {\begin{itemize}%
    \setlength{\itemsep}{0pt}%
    \setlength{\parskip}{0pt}}%
  {\end{itemize}}
  \newenvironment{enumerate*}%
  {\begin{enumerate}%
    \setlength{\itemsep}{0pt}%
    \setlength{\parskip}{0pt}}%
  {\end{enumerate}}

\def\showcomments{1}
\if\showcomments1
\newcommand{\pfliu}[1]{\textcolor{myblue}{\bf\small [#1 --pfliu]}}
\newcommand{\mb}[1]{\textcolor{red}{\bf\small [#1 --mb]}}
\newcommand{\pg}[1]{\textcolor{myyellow}{\bf\small [#1 --pg]}}

\newcommand{\gn}[1]{\textcolor{magenta}{\bf\small [#1 --GN]}}
\else 
\newcommand{\pfliu}[1]{}
\newcommand{\mb}[1]{}
\newcommand{\pg}[1]{}
\newcommand{\gn}[1]{} 
\fi

\graphicspath{./images/}

\title{Re-evaluating Evaluation in Text Summarization
}

\author{Manik Bhandari,
  Pranav Gour,
  Atabak Ashfaq,
  Pengfei Liu,
  Graham Neubig \\
    Carnegie Mellon University \\
    \{\texttt{mbhandar,pgour,aashfaq,pliu3,gneubig\}@cs.cmu.edu}
  }

\date{}

\begin{document}
\maketitle
\begin{abstract}
Automated evaluation metrics as a stand-in for manual evaluation are an essential part of the development of text-generation tasks such as text summarization.
However, while the field has progressed, our standard metrics have not -- for nearly 20 years ROUGE has been the standard evaluation in most summarization papers.
In this paper, we make an attempt to \emph{re-evaluate the evaluation method} for text summarization: assessing the reliability of automatic metrics using \emph{top-scoring system outputs}, both abstractive and extractive, on \emph{recently popular datasets} for both system-level and summary-level evaluation settings. We find that conclusions about evaluation metrics on older datasets do not necessarily hold on modern datasets and systems.
We release a dataset of human judgments that are collected from 25 top-scoring neural summarization systems (14 abstractive and 11 extractive): \url{https://github.com/neulab/REALSumm}
\end{abstract}

\section{Introduction}
\label{sec:introduction}

\begin{table*}[ht]
    \centering
    \small
    \resizebox{\linewidth}{!}{
    \begin{tabular}{lll}
        \toprule
        \textbf{Ability of metrics to} & \textbf{Observations on existing human judgments (\texttt{TAC})} & \textbf{Observations on new human judgments (\texttt{CNNDM})}  \\
        \midrule
        Exp-I: evaluate all systems? (Sec. \ref{sec:williams_test})
        & 
        \begin{minipage}[t]{\columnwidth}
        MoverScore and JS-2 outperform all other metrics.
        \end{minipage}
        &
        \begin{minipage}[t]{\columnwidth}
        ROUGE-2 outperforms all other metrics. MoverScore and JS-2 performs worse both in extractive (only achieved nearly 0.1 \textit{Pearson} correlation) and  abstractive summaries. 
        \end{minipage} \\
        \midrule
        Exp-II: evaluate top-$k$ systems? (Sec. \ref{sec:topk_human})
        & 
        \begin{minipage}[t]{\columnwidth}
        As $k$ becomes smaller, ROUGE-2 de-correlates with humans.
        \end{minipage}
        & 
        \begin{minipage}[t]{\columnwidth}
        For extractive and abstractive systems, ROUGE-2 highly correlates with humans. For evaluating a mix of extractive and abstractive systems, all metrics de-correlate.
        \end{minipage} \\
        \midrule
        Exp-III: compare 2 systems? (Sec. \ref{sec:bootstrap})
        &
        \begin{minipage}[t]{\columnwidth}
        MoverScore and JS-2 outperform all other metrics.
        \end{minipage}
        &
        \begin{minipage}[t]{\columnwidth}
        ROUGE-2 is the most reliable for abstractive systems while ROUGE-1 is most reliable for extractive systems.
        \end{minipage} \\
        \midrule
        Exp-IV: evaluate summaries? (Sec. \ref{sec:summ_level_human}) & 
        \begin{minipage}[t]{\columnwidth}
        (1) MoverScore and JS-2 outperform all other metrics.
        (2) Metrics have much lower correlations when evaluating summaries than systems.
        \end{minipage}
        & 
        \begin{minipage}[t]{\columnwidth}
        (1) ROUGE metrics outperform all other metrics.
        (2) For extractive summaries, most metrics are better at evaluating summaries than systems. For abstractive summaries, some metrics are better at summary level, others are better at system level.
        \end{minipage}  \\
        \bottomrule
    \end{tabular}
    }
    \caption{\label{tab:expt_summary} Summary of our experiments, observations on existing human judgments on the \texttt{TAC}, and contrasting observations on newly obtained human judgments on the \texttt{CNNDM} dataset. Please refer to Sec.~\ref{sec:experiments} for more details.}
\end{table*}

In text summarization, \emph{manual evaluation}, as exemplified by the Pyramid method \cite{nenkova-passonneau-2004-pyramid-og}, is the gold-standard in evaluation.
However, due to time required and relatively high cost of annotation, 
the great majority of research papers on summarization use exclusively automatic evaluation metrics, such as ROUGE \cite{lin2004rouge}
, JS-2~\cite{js2}, S3~\cite{peyrard_s3}, BERTScore~\cite{bert-score}, MoverScore~\cite{zhao-etal-2019-moverscore} etc.
Among these metrics, ROUGE is by far the most popular, and there is relatively little discussion of how ROUGE may deviate from human judgment and the potential for this deviation to change conclusions drawn regarding relative merit of baseline and proposed methods.
To characterize the relative goodness of evaluation metrics, it is necessary to perform \emph{meta-evaluation} \cite{graham-2015-evaluating,lin-och-2004-orange}, where a dataset annotated with human judgments (e.g.~TAC\footnote{\href{https://tac.nist.gov/}{https://tac.nist.gov/}} 2008~\cite{tac2008}) is used to test the degree to which automatic metrics correlate therewith.


However, the classic TAC meta-evaluation datasets are now 6-12 years old\footnote{In TAC, summarization was in 2008, 2009, 2010, 2011, 2014. In 2014, the task was biomedical summarization.} and it is not clear whether conclusions found there will hold with modern systems and summarization tasks. Two earlier works exemplify this disconnect: (1) \citet{peyrard-2019-studying} observed that the human-annotated summaries in the TAC dataset are mostly of lower quality than those produced by modern systems and that various automated evaluation metrics strongly disagree in the higher-scoring range in which current systems now operate. (2) \citet{rankel-etal-2013-decade} observed that the correlation between ROUGE and human judgments in the TAC dataset decreases when looking at the best systems only, even for systems from eight years ago, which are far from today's state-of-the-art.

Constrained by few existing human judgment datasets, it remains unknown how existing metrics behave on current top-scoring summarization systems.
In this paper, we ask the question: does the rapid progress of model development in summarization models require us to \emph{re-evaluate} the evaluation process used for text summarization?
To this end, we create and release a large benchmark for meta-evaluating summarization metrics including:%




\begin{itemize*}
    \item \textbf{Outputs} from 25 top-scoring extractive and abstractive summarization systems on the \texttt{CNN/DailyMail} dataset.
    \item \textbf{Automatic evaluations} from several evaluation metrics including traditional metrics (e.g.~ROUGE) and modern semantic matching metrics (e.g.~BERTScore, MoverScore).
    \item \textbf{Manual evaluations} using the lightweight pyramids method~\cite{litepyramids-shapira-etal-2019-crowdsourcing}, which we use as a gold-standard to evaluate summarization systems as well as automated metrics.
\end{itemize*}


Using this benchmark, we perform an extensive analysis, which indicates the need to re-examine our assumptions about the evaluation of automatic summarization systems.
Specifically, we conduct four experiments analyzing the correspondence between various metrics and human evaluation.
Somewhat surprisingly, we find that many of the previously attested properties of metrics found on the \texttt{TAC} dataset demonstrate different trends on our newly collected \texttt{CNNDM} dataset, as shown in Tab.~\ref{tab:expt_summary}.
For example, MoverScore is the best performing metric for evaluating summaries on dataset \texttt{TAC}, but it is significantly worse than ROUGE-2 on our collected \texttt{CNNDM} set.
Additionally, many previous works~\cite{novikova-etal-2017-need,peyrard_s3,chaganty2018price} show that metrics have much lower correlations at comparing summaries than systems.
For extractive summaries on \texttt{CNNDM}, however, most metrics are better at comparing summaries than systems. 

\paragraph{Calls for Future Research}
These observations demonstrate the limitations of our current best-performing metrics, highlighting
(1) the need for future meta-evaluation to (i) be across multiple datasets and (ii) evaluate metrics on different application scenarios, e.g.~summary level vs.~system level
(2) the need for more systematic meta-evaluation of summarization metrics that updates with our ever-evolving systems and datasets, and
(3) the potential benefit to the summarization community of a shared task similar to the WMT\footnote{http://www.statmt.org/wmt20/} Metrics Task in Machine Translation, where systems and metrics co-evolve.

\section{Preliminaries}

In this section we describe the datasets, systems, metrics, and meta evaluation methods used below.

\subsection{Datasets}  

\noindent\textbf{TAC-2008, 2009}~\cite{tac2008,tac2009} are multi-document, multi-reference summarization datasets. Human judgments are available on for the system summaries submitted during the TAC-2008, TAC-2009 shared tasks.

\noindent\textbf{CNN/DailyMail (CNNDM)} \cite{hermann2015teaching,nallapati2016abstractive} is a commonly used summarization dataset that contains news articles and
associated highlights as summaries. We use the version without entities anonymized. 

\subsection{Representative Systems} \label{sec:system}


We use the following representative top-scoring systems that either achieve state-of-the-art (SOTA) results or competitive performance, for which we could gather the outputs on the \texttt{CNNDM} dataset.

\noindent \textbf{Extractive summarization systems.}
We use  
CNN-LSTM-BiClassifier (CLSTM-SL; \citet{kedzie2018content}),
Latent~\cite{zhang-etal-2018-neural-latent},
BanditSum~\cite{dong-etal-2018-banditsum}, 
REFRESH~\cite{narayan-etal-2018-ranking},
NeuSum \cite{zhou-etal-2018-neural}, 
HIBERT ~\cite{zhang-etal-2019-hibert},
Bert-Sum-Ext~\cite{liu-lapata-2019-text},
CNN-Transformer-BiClassifier (CTrans-SL; \citet{zhong2019searching}), 
CNN-Transformer-Pointer (CTrans-PN; \citet{zhong2019searching}),
HeterGraph~\cite{wang2020heterogeneous} and MatchSum~\cite{zhong2020extractive} 
as representatives of extractive systems, totaling 11 extractive system outputs for each document in the \texttt{CNNDM} test set. 

\noindent \textbf{Abstractive summarization systems.}
We use pointer-generator+coverage \cite{see-etal-2017-get}, 
fastAbsRL \cite{chen-bansal-2018-fast-abs}, 
fastAbsRL-rank \cite{chen-bansal-2018-fast-abs}, 
Bottom-up \cite{gehrmann2018bottom}, 
T5 \cite{raffel2019exploring-t5}, 
Unilm-v1 \cite{dong2019unified}, 
Unilm-v2 \cite{dong2019unified}, 
twoStageRL \cite{zhang2019pretraining}, 
preSummAbs \cite{liu2019text-presumm},
preSummAbs-ext \cite{liu2019text-presumm} 
BART \cite{lewis2019bart} and
Semsim \cite{yoon2020learning}
as abstractive systems. In total, we use 14 abstractive system outputs for each document in the \texttt{CNNDM} test set. 
\subsection{Evaluation Metrics}
\label{sec:metrics}
We examine eight metrics that measure the agreement between two texts, in our case, between the system summary and reference summary.  


\noindent\textbf{BERTScore (BScore)} measures soft overlap between contextual BERT embeddings of tokens between the two texts\footnote{Used code at \href{https://github.com/Tiiiger/bert_score}{github.com/Tiiiger/bert\_score}} \citep{bert-score}.

\noindent\textbf{MoverScore (MScore)} applies a distance measure to contextualized BERT and ELMo word embeddings\footnote{Used code at \href{https://github.com/AIPHES/emnlp19-moverscore}{github.com/AIPHES/emnlp19-moverscore}} \citep{zhao-etal-2019-moverscore}.

\noindent\textbf{Sentence Mover Similarity (SMS)} applies minimum distance matching between text based on sentence embeddings \citep{clark2019sentence}.

\noindent\textbf{Word Mover Similarity (WMS)} measures similarity using minimum distance matching between texts which are represented as a bag of word embeddings\footnote{For WMS and SMS: \href{https://github.com/eaclark07/sms}{github.com/eaclark07/sms}} \citep{kusner2015word}.

\noindent\textbf{JS divergence (JS-2)} measures Jensen-Shannon divergence between the two text's bigram distributions\footnote{JS-2 is calculated using the function defined in \href{https://github.com/UKPLab/coling2016-genetic-swarm-MDS}{github.com/UKPLab/coling2016-genetic-swarm-MDS}} \citep{lin-etal-2006-information}. 

\noindent\textbf{ROUGE-1 and ROUGE-2}\ measure overlap of unigrams and bigrams respectively\footnote{\label{pyrouge}For ROUGE-1,2, and L, we used the python wrapper: \href{https://github.com/sebastianGehrmann/rouge-baselines}{https://github.com/sebastianGehrmann/rouge-baselines}} \citep{lin2004rouge}.

\noindent\textbf{ROUGE-L} measures overlap of the longest common subsequence between two texts \citep{lin2004rouge}.

\noindent
We use the recall variant of all metrics (since the Pyramid method of human evaluations is inherently recall based) except MScore which has no specific recall variant.



\subsection{Correlation Measures}



\noindent\textbf{\textit{Pearson} Correlation}
is a measure of linear correlation between two variables and is popular in meta-evaluating metrics at the system level \citep{lee1988thirteen}. We use the implementation given by~\citet{scipy-2020}.

\noindent\textbf{William's Significance Test} is a means of calculating the statistical significance of differences in correlations for dependent variables \cite{williams-test,graham-baldwin-2014-testing}. This is useful for us since metrics evaluated on the same dataset are not independent of each other.

\subsection{Meta Evaluation Strategies}
\label{sec:meta_eval_strategy}
There are two broad meta-evaluation strategies: summary-level and system-level.

\noindent \textbf{Setup:} For each document $d_i, i \in \{1 \dots n\}$ in a dataset $\mathcal{D}$, we have $J$ system outputs, where the outputs can come from (1) extractive systems (Ext), (2) abstractive systems (Abs) or (3) a union of both (Mix). Let $s_{ij}, j \in \{1 \dots J\}$ be the $j^{th}$ summary of the $i^{th}$ document, $m_i$ be a specific metric and $K$ be a correlation measure.

\subsubsection{Summary Level}
Summary-level correlation is calculated as follows:
\begin{equation}
    \begin{split}
        K_{m_1 m_2}^{sum} &= \frac{1}{n} \sum_{i = 1}^{n} \bigg( K\big([m_1(s_{i1}) \dots m_1(s_{iJ})], \\
        & ~~~~~~~~~~~~~~~~~~~~~~~~[m_2(s_{i1}) \dots m_2(s_{iJ}) ]\big) \bigg).
    \end{split}
    \label{eqn:summ_level_corr}
\end{equation}
Here, correlation is calculated for each document, among the different system outputs of that document, and the mean value is reported.


\subsubsection{System Level}
System-level correlation is calculated as follows:
\begin{footnotesize}
        \begin{align}
        K_{m_1 m_2}^{sys} = K\bigg( &\left[\frac{1}{n}\sum_{i = 1}^n m_1(s_{i1}) \dots \frac{1}{n}\sum_{i = 1}^n m_1(s_{iJ})\right], \notag \\
        & \left[\frac{1}{n}\sum_{i = 1}^n m_2(s_{i1}) \dots \frac{1}{n}\sum_{i = 1}^n m_2(s_{iJ})\right]\bigg).
    \end{align}
\end{footnotesize}
Additionally, the ``quality" of a system $sys_j$ is defined as the mean human score received by it i.e. 
\begin{equation}
    \mathrm{HScore}_{mean}^{sys_j} = \frac{1}{n}\sum_{i=1}^{n} \mathrm{humanScore}(s_{ij}).
    \label{eqn:human_score}
\end{equation}


\section{Collection of Human Judgments}
\label{sec:human_collection}
We follow a 3-step process to collect human judgments:
(1) we \emph{collect} system-generated summaries on the most-commonly used summarization dataset, \texttt{CNNDM};  
(2) we \emph{select} representative test samples from \texttt{CNNDM} and 
(3) we manually \emph{evaluate} system-generated summaries of the above-selected test samples.

\subsection{System-Generated Summary Collection}
We collect the system-generated summaries from 25 top-scoring systems,\footnote{We contacted the authors of these systems to gather the corresponding outputs, including variants of the systems.} covering 11 extractive and 14 abstractive systems (Sec.~\ref{sec:system}) on the \texttt{CNNDM} dataset. We organize our collected generated summaries into three groups based on system type:

\begin{itemize*}
    \item \texttt{CNNDM Abs} denotes collected output summaries from abstractive systems.
    \item \texttt{CNNDM Ext} denotes collected output summaries from extractive systems.
    \item \texttt{CNNDM Mix} is the union of the two.
\end{itemize*}

\subsection{Representative Sample Selection}


Since collecting human annotations is costly, we sample 100 documents from \texttt{CNNDM} test set (11,490 samples) and evaluate system generated summaries of these 100 documents. We aim to include documents of varying difficulties in the representative sample. As a proxy to the difficulty of summarizing a document, we use the mean score received by the system generated summaries for the document. Based on this, we partition the \texttt{CNNDM} test set into $5$ equal sized bins and sample 4 documents from each bin.
We repeat this process for 5 metrics (BERTScore, MoverScore, R-1, R-2, R-L) obtaining a sample of 100 documents. This methodology is detailed in Alg.~\ref{alg:sampling} in Sec. A.1. 



\begin{table*}[ht]
    \centering
    \small
    {
    \begin{tabular}{c}
        \toprule
        \begin{minipage}[t]{0.98\linewidth}
        \textbf{(a) Reference Summary}: 
        Bayern Munich beat Porto 6 - 1 in the Champions League on Tuesday. Pep Guardiola's side progressed 7 - 4 on aggregate to reach semi-finals. Thomas Muller scored 27th Champions League goal to pass Mario Gomez. Muller is now the leading German scorer in the competition. After game Muller led the celebrations with supporters using a megaphone. \\
        \end{minipage}\tabularnewline
        \begin{minipage}[t]{0.98\linewidth}
        \textbf{(b) System Summary (BART, \citet{lewis2019bart}):} 
        Bayern Munich beat Porto 6 - 1 at the Allianz Arena on Tuesday night. Thomas Muller scored his 27th Champions League goal. The 25 - year - old became the highest - scoring German since the tournament took its current shape in 1992. Bayern players remained on the pitch for some time as they celebrated with supporters. \\
        \end{minipage}\tabularnewline 
        \begin{minipage}[t]{0.98\linewidth}
        \textbf{(c) SCUs with corresponding evaluations}:  
        \vspace{-2\topsep}
        \begin{multicols}{2}
        \begin{itemize*}
            \item Bayern Munich beat Porto.~$\checkmark$
            \item Bayern Munich won  6 - 1.~$\checkmark$
            \item Bayern Munich won in Champions League.~$\checkmark$
            \item Bayern Munich won on Tuesday.~$\checkmark$
            \item Bayern Munich is managed by Pep Guardiola.~$\times$
            \item Bayern Munich progressed in the competition.~$\checkmark$
            \item Bayern Munich reached semi-finals.~$\times$
            \item Bayern Munich progressed 7 - 4 on aggregate.~$\times$
            \item Thomas Muller scored 27th Champions League goal.~$\checkmark$
            \item Thomas Muller passed Mario Gomez in goals.~$\times$
            \item Thomas Muller is now the leading German scorer in the competition.~$\checkmark$
            \item After the game Thomas Muller led the celebrations.~$\times$
            \item Thomas Muller led the celebrations using a megaphone.~$\times$
    \end{itemize*}  
    \end{multicols}
    \end{minipage}\tabularnewline \\
    \bottomrule
    \end{tabular}
    }
    \caption{\label{tab:example_annotation}Example of a summary and corresponding annotation. (a) shows a reference summary from the representative sample of the \texttt{CNNDM} test set. (b) shows the corresponding system summary generated by BART, one of the abstractive systems used in the study. (c) shows the SCUs (Semantic Content Units) extracted from (a) and the ``\textit{Present}($\checkmark$)''/``\textit{Not Present}($\times$)'' marked by crowd workers when evaluating (b). 
    }
\end{table*}

\subsection{Human Evaluation}


In text summarization, a ``good'' summary should represent as much relevant content from the input document as possible, within the acceptable length limits.
Many human evaluation methods have been proposed to capture this desideratum \cite{nenkova-passonneau-2004-pyramid-og,chaganty2018price,fan-etal-2018-controllable,litepyramids-shapira-etal-2019-crowdsourcing}.
Among these,  \emph{Pyramid} \cite{nenkova-passonneau-2004-pyramid-og} is a reliable and widely used method, that evaluates content selection by (1) exhaustively obtaining Semantic Content Units (SCUs) from reference summaries, (2) weighting them based on the number of times they are mentioned and (3) scoring a system summary based on which SCUs can be inferred. 

Recently, \citet{litepyramids-shapira-etal-2019-crowdsourcing} extended Pyramid to a lightweight, crowdsourcable method - LitePyramids, which uses Amazon Mechanical Turk\footnote{\href{https://www.mturk.com/}{https://www.mturk.com/}} (AMT) for gathering human annotations. LitePyramids simplifies Pyramid by (1) allowing crowd workers to extract a subset of all possible SCUs and (2) eliminating the difficult task of merging duplicate SCUs from different reference summaries, instead using SCU sampling to simulate frequency-based weighting. 

Both Pyramid and LitePyramid rely on the presence of multiple references per document to assign importance weights to SCUs. However in the \texttt{CNNDM} dataset there is only one reference summary per document. We therefore adapt the LitePyramid method for the single-reference setting as follows.





\noindent \textbf{SCU Extraction} The LitePyramids annotation instructions define a Semantic Content Unit (SCU) as a \textit{sentence containing a single fact written as briefly and clearly as possible}.
Instead, we focus on shorter, more fine-grained SCUs that contain at most 2-3 entities. This allows for partial content overlap between a generated and reference summary, and also makes the task easy for workers. Tab.~\ref{tab:example_annotation} gives an example. We exhaustively extract (up to 16) SCUs\footnote{In our representative sample we found no document having more than 16 SCUs.} from each reference summary.
Requiring the set of SCUs to be exhaustive increases the complexity of the SCU generation task, and hence instead of relying on crowd-workers, we create SCUs from reference summaries ourselves.
In the end, we obtained nearly 10.5 SCUs on average from each reference summary.

\noindent \textbf{System Evaluation} During system evaluation the full set of SCUs is presented to crowd workers. Workers are paid similar to~\citet{litepyramids-shapira-etal-2019-crowdsourcing}, scaling the rates for fewer SCUs and shorter summary texts. For abstractive systems, we pay \$0.20 per summary and for extractive systems, we pay \$0.15 per summary since extractive summaries are more readable and might precisely overlap with SCUs. 
We post-process system output summaries before presenting them to annotators by true-casing the text using Stanford CoreNLP \cite{manning-EtAl-2014-CoreNLP} and replacing ``unknown'' tokens with a special symbol ``$\Box$'' \citep{chaganty2018price}.

Tab.~\ref{tab:example_annotation} depicts an example reference summary, system summary, SCUs extracted from the reference summary, and annotations obtained in evaluating the system summary.

\noindent \textbf{Annotation Scoring}
For robustness~\cite{litepyramids-shapira-etal-2019-crowdsourcing}, each system summary is evaluated by 4 crowd workers. Each worker annotates up to 16 SCUs by marking an SCU ``present" if it can be inferred from the system summary or ``not present" otherwise. We obtain a total of 10,000 human annotations (100 documents~$\times$ 25 systems~$\times$ 4 workers). For each document, we identify a ``noisy" worker as one who disagrees with the majority (i.e. marks an SCU as ``present" when majority thinks ``not present" or vice-versa), on the largest number of SCUs. We remove the annotations of noisy workers and retain 7,742 annotations of the 10,000. After this filtering, we obtain an average inter-annotator agreement (Krippendorff's alpha~\cite{krippendorff2011computing}) of 0.66.%
\footnote{The agreement was 0.57 and 0.72 for extractive and abstractive systems respectively.}
Finally, we use the majority vote to mark the presence of an SCU in a system summary, breaking ties by the class, ``not present".

\section{Experiments}
\label{sec:experiments}
Motivated by the central research question: ``does the rapid progress of model development in summarization models require us to \emph{re-evaluate} the evaluation process used for text summarization?''
We use the collected human judgments 
to meta-evaluate current metrics from four diverse viewpoints, measuring the ability of metrics to: (1) evaluate \textit{all} systems; (2) evaluate top-$k$ strongest systems; (3) compare \textit{two} systems; (4) evaluate individual summaries. 
We find that many previously attested properties of metrics observed on \texttt{TAC} exhibit different trends on the new \texttt{CNNDM} dataset.


\subsection{Exp-I: Evaluating \textit{All} Systems}
\label{sec:williams_test}

\begin{figure*}[!t]
  \centering
  \subfloat[TAC-2008]{
    \label{fig:tac_8_william}
    \includegraphics[width=0.2225\linewidth]{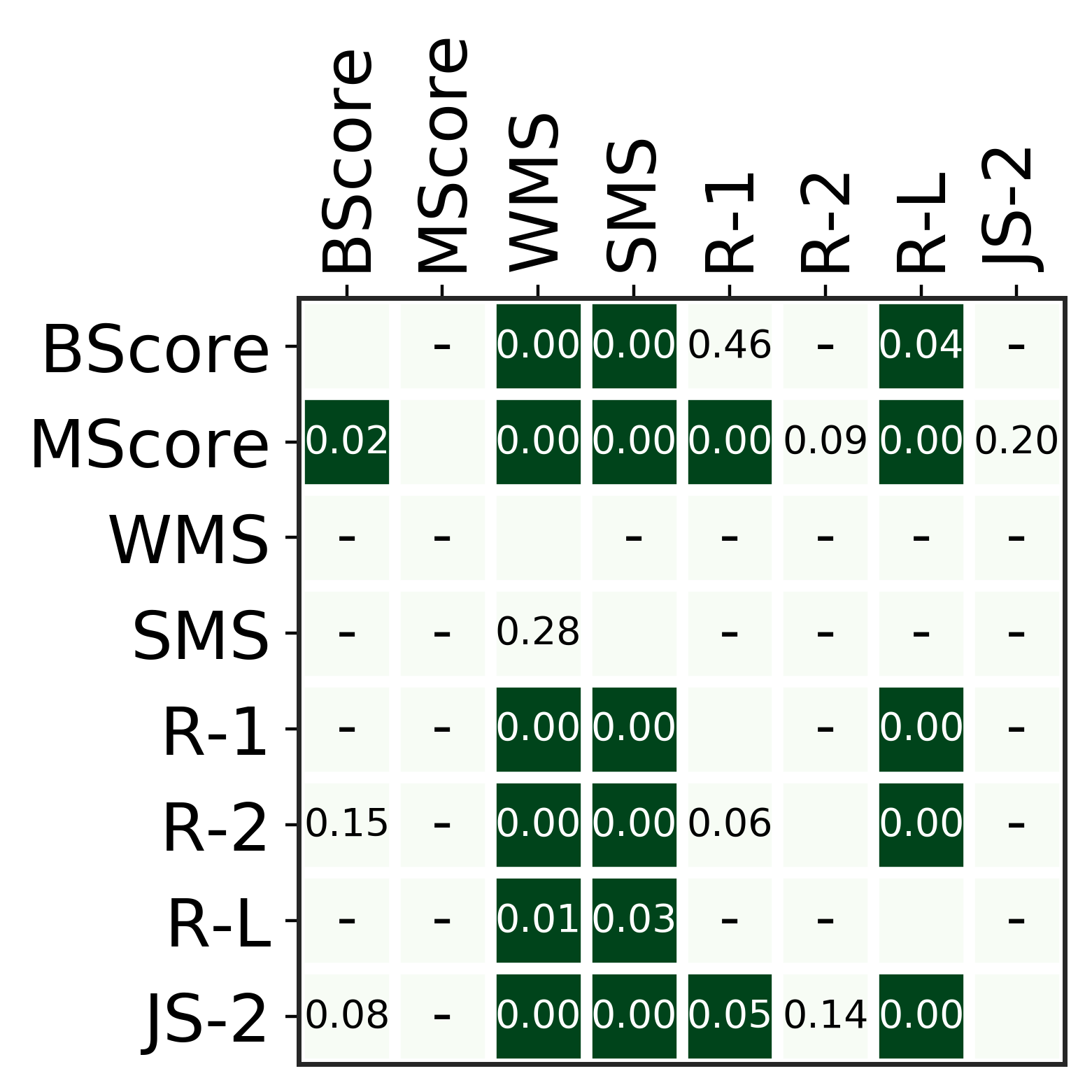}
  }
  \subfloat[TAC-2009]{
    \label{fig:tac_9_william}
    \includegraphics[width=0.17\linewidth]{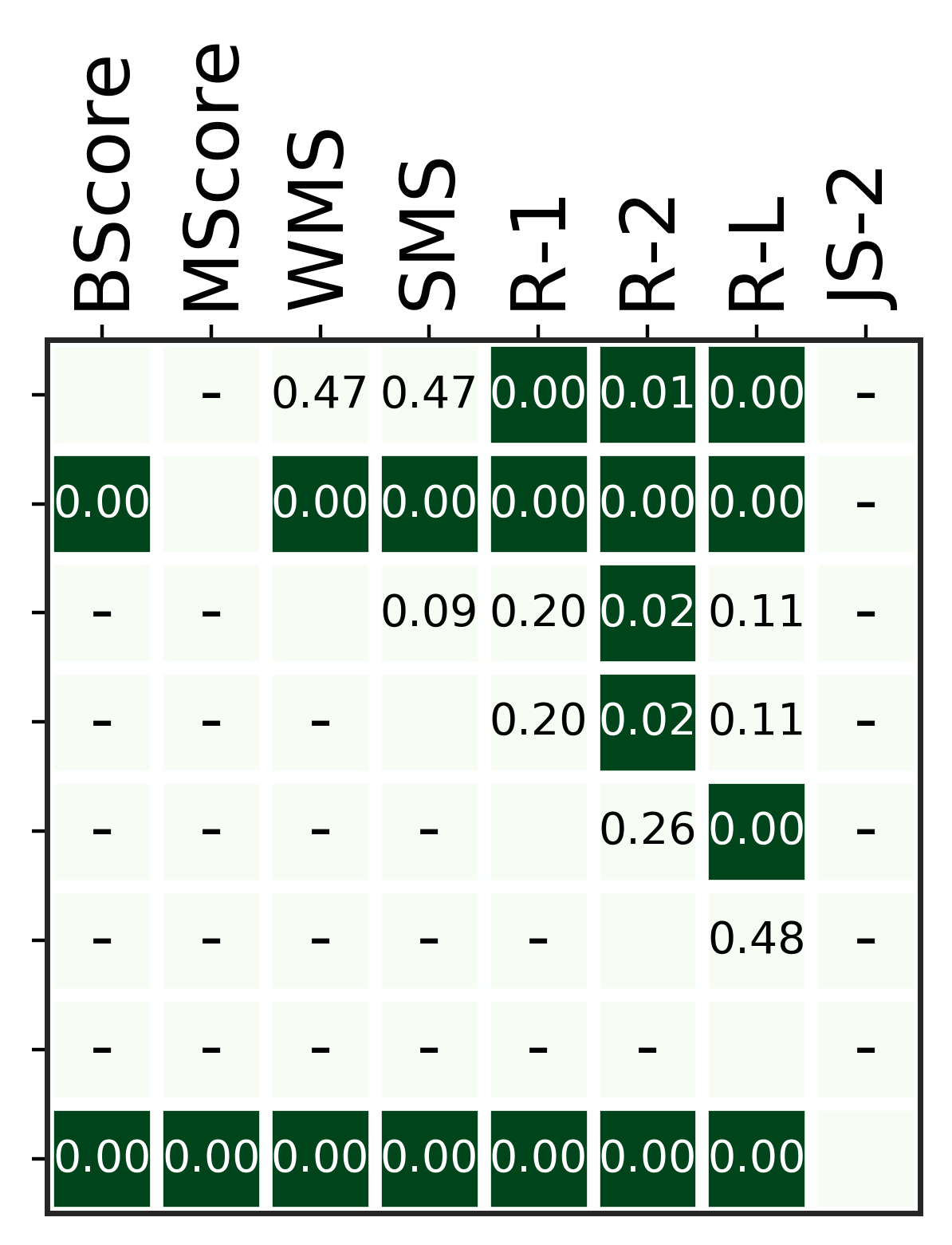}
  }
  \subfloat[CNNDM Mix]{
    \label{fig:cnn_dm_mix_william}
    \includegraphics[width=0.17\linewidth]{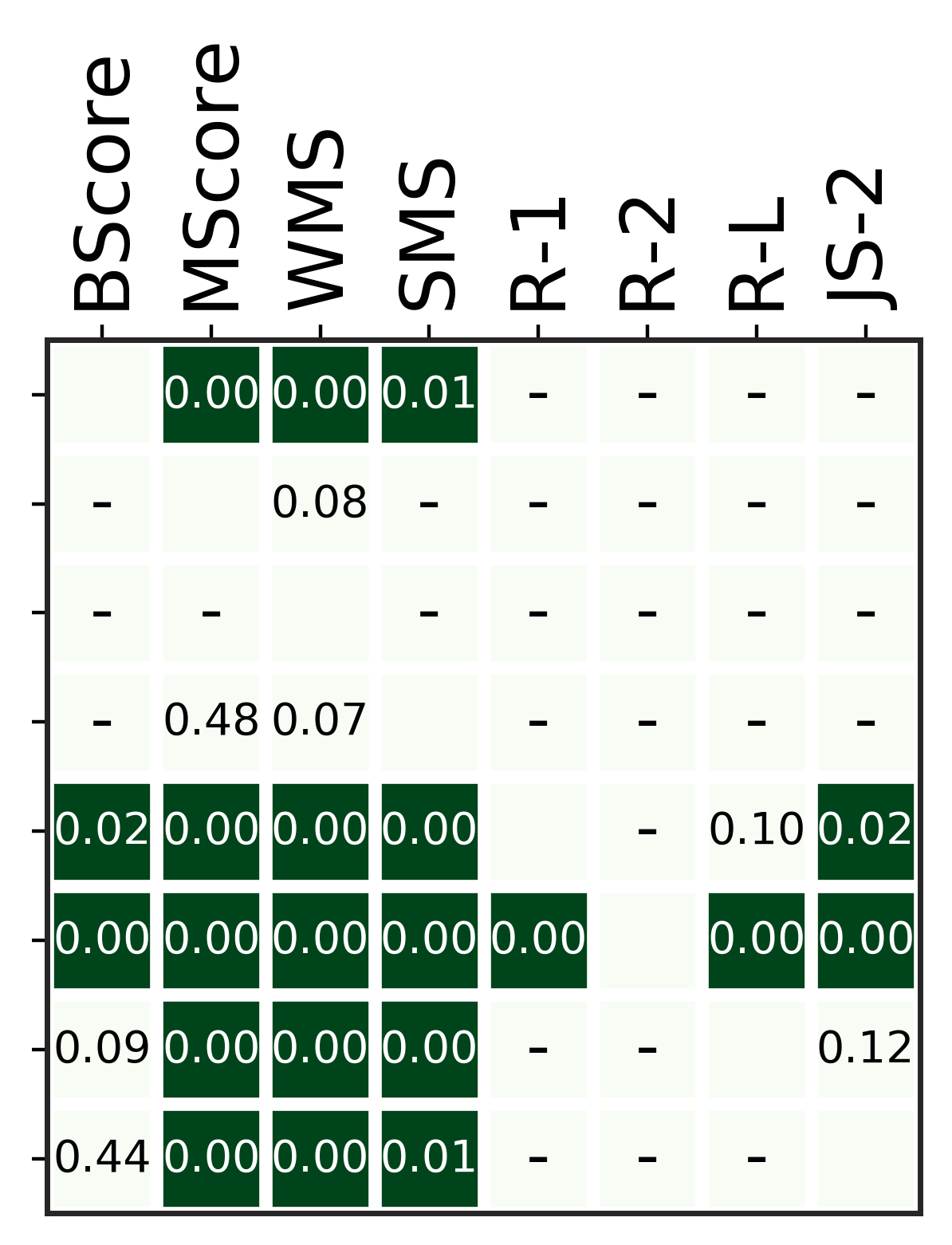}
  }
  \subfloat[CNNDM Abs]{
    \label{fig:cnn_dm_abs_william}
    \includegraphics[width=0.17\linewidth]{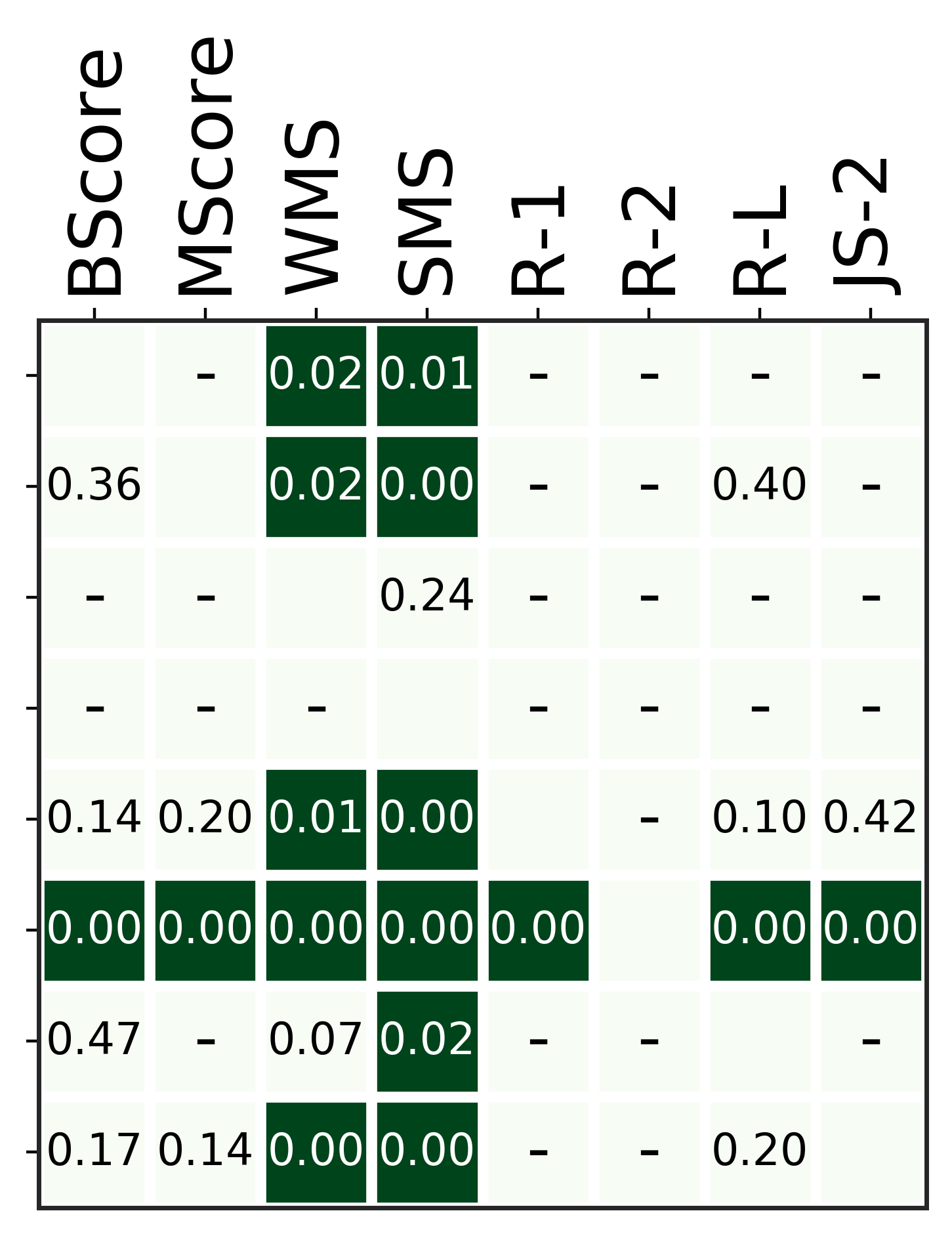}
  }
  \subfloat[CNNDM Ext]{
    \label{fig:cnn_dm_ext_william}
    \includegraphics[width=0.17\linewidth]{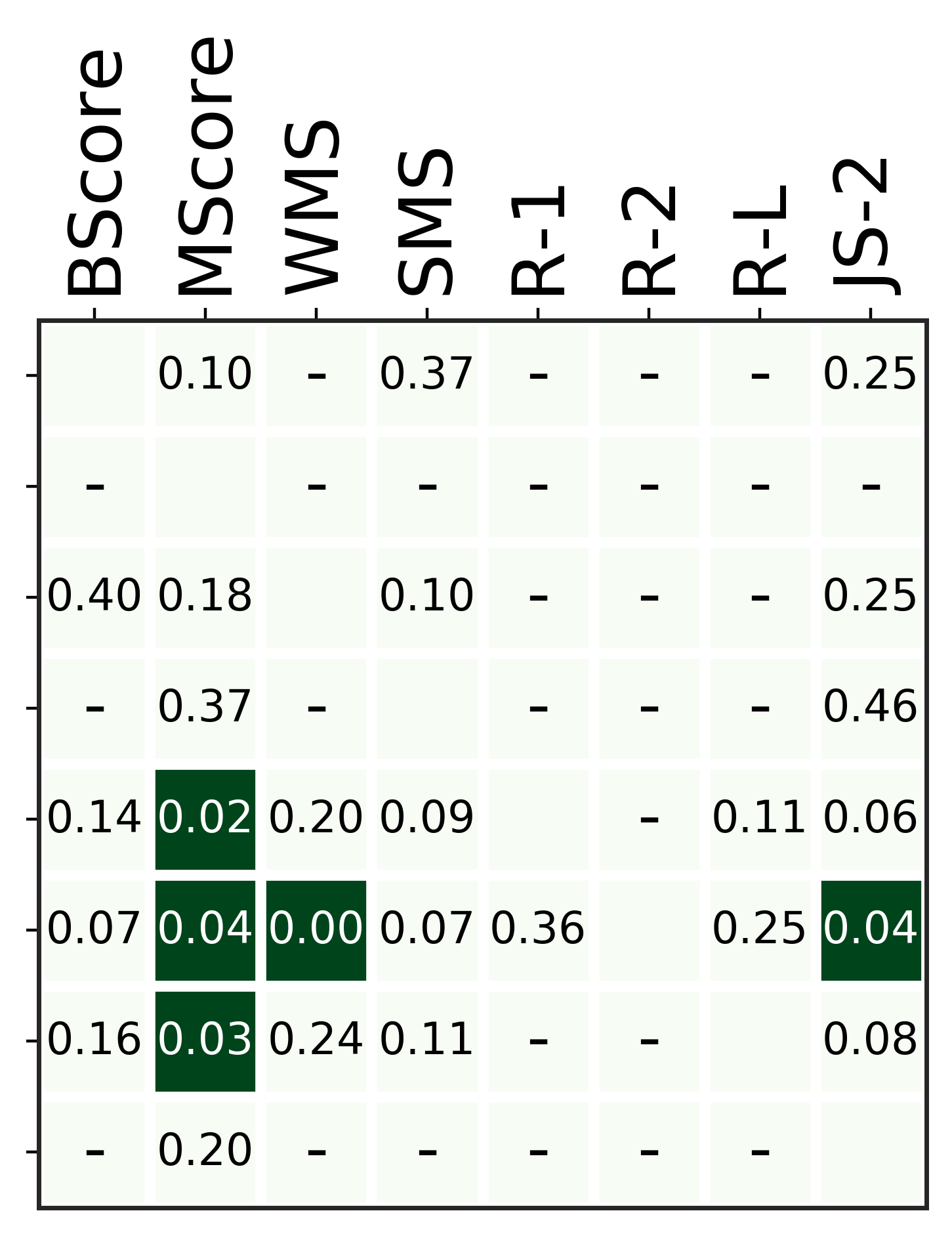}
  }

\caption{\label{fig:williams_test}p-value of William's Significance Test for the hypothesis ``Is the system on left (y-axis) significantly better than system on top (x-axis)". `BScore' refers to BERTScore and `MScore' refers to MoverScore. A dark green value in cell $(i, j)$ denotes metric $m_i$ has a significantly higher \textit{Pearson} correlation with human scores compared to metric $m_j$ (p-value $< 0.05$).\protect\footnotemark `-' in cell $(i, j)$ refers to the case when \textit{Pearson} correlation of $m_i$ with human scores is less that of $m_j$ (Sec.~\ref{sec:williams_test}).
}
\end{figure*}
\footnotetext{{Dark cells with p-value $= 0.05$ have been rounded up.}}

\begin{figure}
    \centering
    \includegraphics[width=\linewidth]{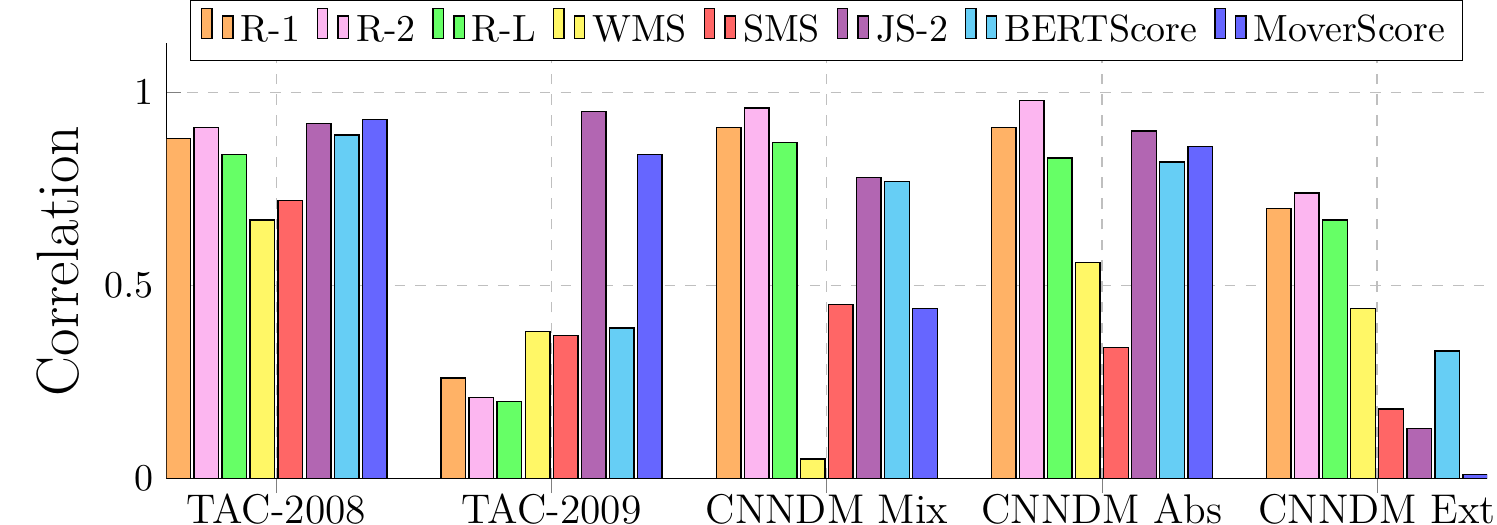}
  
    \caption{\label{fig:pearson-mix}System-level \textit{Pearson} correlation between metrics and human scores (Sec.~\ref{sec:williams_test}).
    }
\end{figure}

Automatic metrics are widely used to determine where a new system may rank against existing state-of-the-art systems. Thus, in meta-evaluation studies, calculating correlation of automatic metrics with human judgments at the system level is a commonly-used setting~\cite{novikova-etal-2017-need, bojar-etal-2016-wmt-results, graham-2015-evaluating}.
We follow this setting 
and specifically, ask two questions:
\noindent \textbf{Can metrics reliably compare different systems?}
To answer this we observe the \textit{Pearson} correlation between different metrics and human judgments in Fig.~\ref{fig:pearson-mix}, finding that: 

\noindent (1) MoverScore and JS-2, which were the best performing metrics on \texttt{TAC}, have poor correlations with humans in comparing \texttt{CNNDM Ext} systems.

\noindent (2) Most metrics have high correlations on the \texttt{TAC-2008} dataset but many suffer on \texttt{TAC-2009}, especially ROUGE based metrics. However, ROUGE metrics consistently perform well on the collected \texttt{CNNDM} datasets.



\noindent \textbf{Are some metrics significantly better than others in comparing systems? }
Since automated metrics calculated on the same data are not independent, 
we must perform the William's test~\cite{williams-test} to establish if the difference in correlations between metrics is statistically significant~\cite{graham-baldwin-2014-testing}. In Fig.~\ref{fig:williams_test} we report the p-values of William's test. 
We find that 

\noindent
(1) MoverScore and JS-2 are significantly better than other metrics in correlating with human judgments on the \texttt{TAC} datasets.

\noindent
(2) However, on \texttt{CNNDM Abs} and \texttt{CNNDM Mix}, R-2 significantly outperforms all others whereas on \texttt{CNNDM Ext} none of the metrics show significant improvements over others.

\noindent \textbf{Takeaway:}
These results suggest that metrics run the risk of overfitting to some datasets, highlighting the need to meta-evaluate metrics for modern datasets and systems.
Additionally, there is no one-size-fits-all metric that can outperform others on all datasets. This suggests the utility of using different metrics for different datasets to evaluate systems e.g. MoverScore on \texttt{TAC-2008}, JS-2 on \texttt{TAC-2009} and R-2 on \texttt{CNNDM} datasets. 
\subsection{Exp-II: Evaluating \textit{Top-k} Systems}
\label{sec:topk_human}
\begin{figure*}[t]
  \centering
  \subfloat[TAC-2008]{
    \label{fig:tac_8_topk}
    \includegraphics[width=0.205\linewidth]{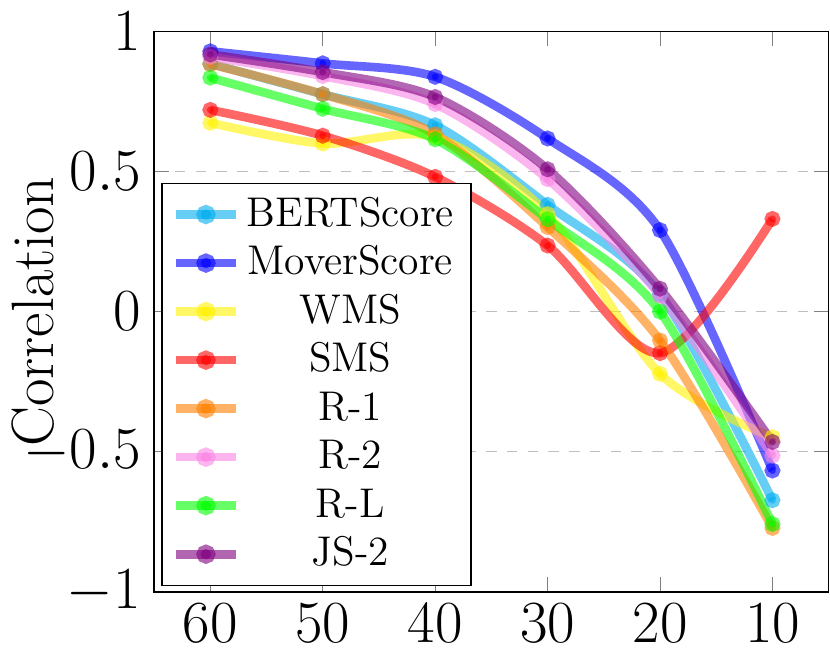}
  }
  \subfloat[TAC-2009]{
    \label{fig:xsum}
    \includegraphics[width=0.1725\linewidth]{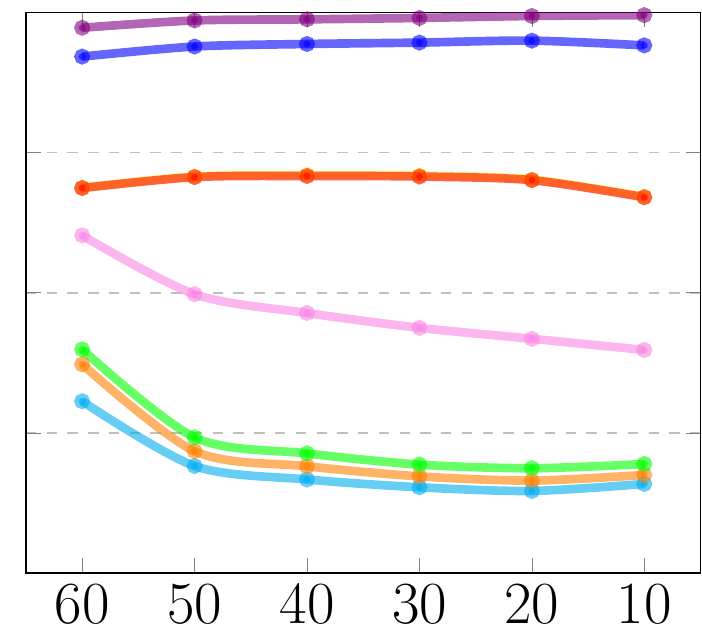}
  }
  \subfloat[CNNDM Mix]{
    \label{fig:cnn_dm_mix_topk}
    \includegraphics[width=0.1725\linewidth]{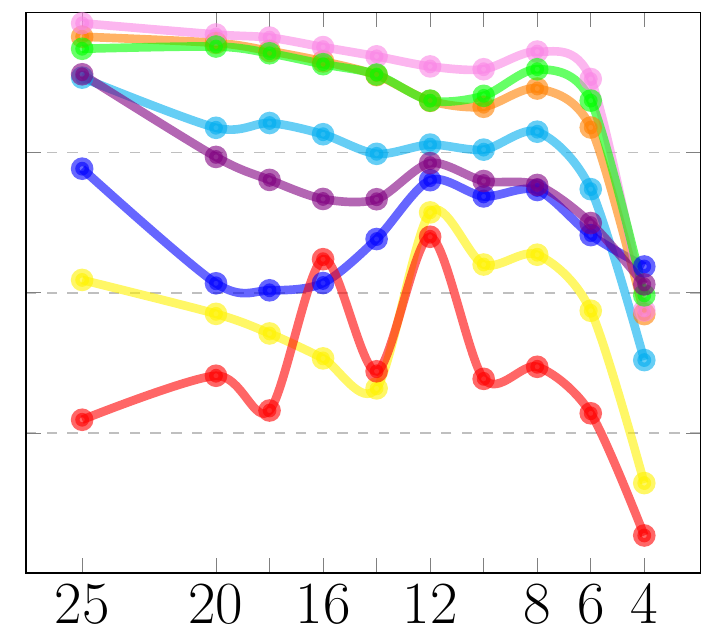}
  }
  \subfloat[CNNDM Abs]{
    \label{fig:cnn_dm_abs_topk}
    \includegraphics[width=0.1725\linewidth]{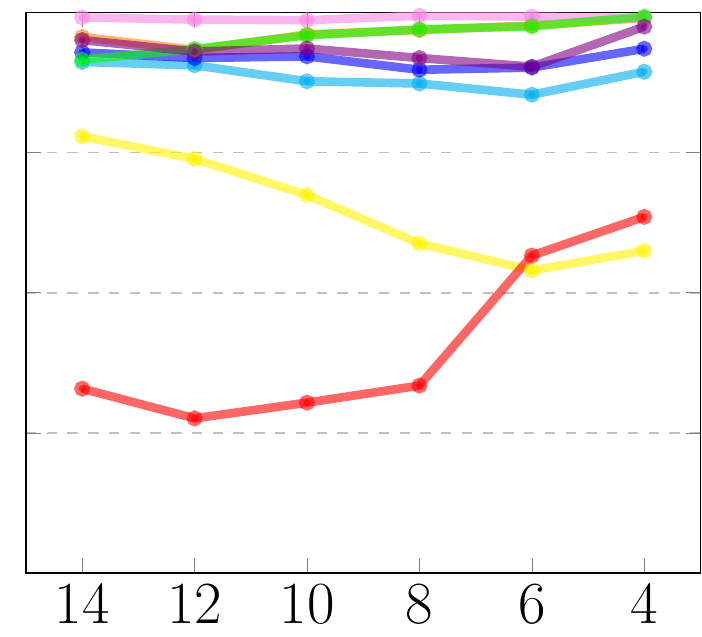}
  }
  \subfloat[CNNDM Ext]{
    \label{fig:cnn_dm_ext_topk}
    \includegraphics[width=0.1725\linewidth]{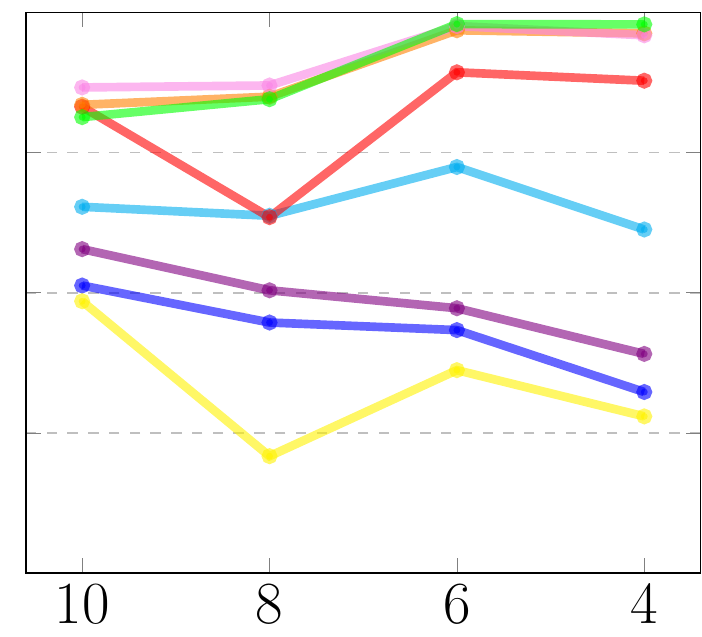}
  }
  
\caption{System-level \textit{Pearson} correlation with humans on top-$k$ systems (Sec.~\ref{sec:topk_human}).
}
 \label{fig:top_k_human}
\end{figure*}

Most papers that propose a new state-of-the-art system often use 
automatic metrics as a proxy to human judgments to compare their proposed method against other top scoring systems. However, \textit{can metrics reliably quantify the improvements that one high quality system makes over other competitive systems?}
To answer this, instead of focusing on all of the collected systems, we evaluate the correlation between automatic metrics and human judgments in comparing the \textit{top-k} systems, where \textit{top-k} are chosen based on a system's mean human score (Eqn.~\ref{eqn:human_score}).%
\footnote{As a caveat, we \emph{do not} perform significance testing for this experiment, due to the small number of data points.}
Our observations are presented in Fig.~\ref{fig:top_k_human}. We find that: 

\noindent
(1) As $k$ becomes smaller, metrics de-correlate with humans on the \texttt{TAC-2008} and \texttt{CNNDM Mix} datasets, even getting negative correlations for small values of $k$ (Fig. \ref{fig:tac_8_topk}, \ref{fig:cnn_dm_mix_topk}). Interestingly, SMS, R-1, R-2 and R-L \textit{improve} in performance as $k$ becomes smaller on \texttt{CNNDM Ext}. 

\noindent
(2) R-2 had negative correlations with human judgments on \texttt{TAC-2009} for $k < 50$, however it remains highly correlated with human judgments on \texttt{CNNDM Abs} for all values of $k$. 


\noindent \textbf{Takeaway:} Metrics cannot reliably quantify the improvements made by one system over others, especially for the top few systems across all datasets. Some metrics, however, are well suited for specific datasets, e.g. JS-2 and R-2 are reliable indicators of improvements on \texttt{TAC-2009} and \texttt{CNNDM Abs} respectively.

\subsection{Exp-III: Comparing $Two$-Systems}
\label{sec:bootstrap}



Instead of comparing many systems (Sec.~\ref{sec:williams_test}, ~\ref{sec:topk_human}) ranking $two$ systems aims to test the discriminative power of a metric, i.e., the degree to which the metric can capture statistically significant differences between two summarization systems.

\begin{figure}
    \centering
    \includegraphics[width=0.95\linewidth]{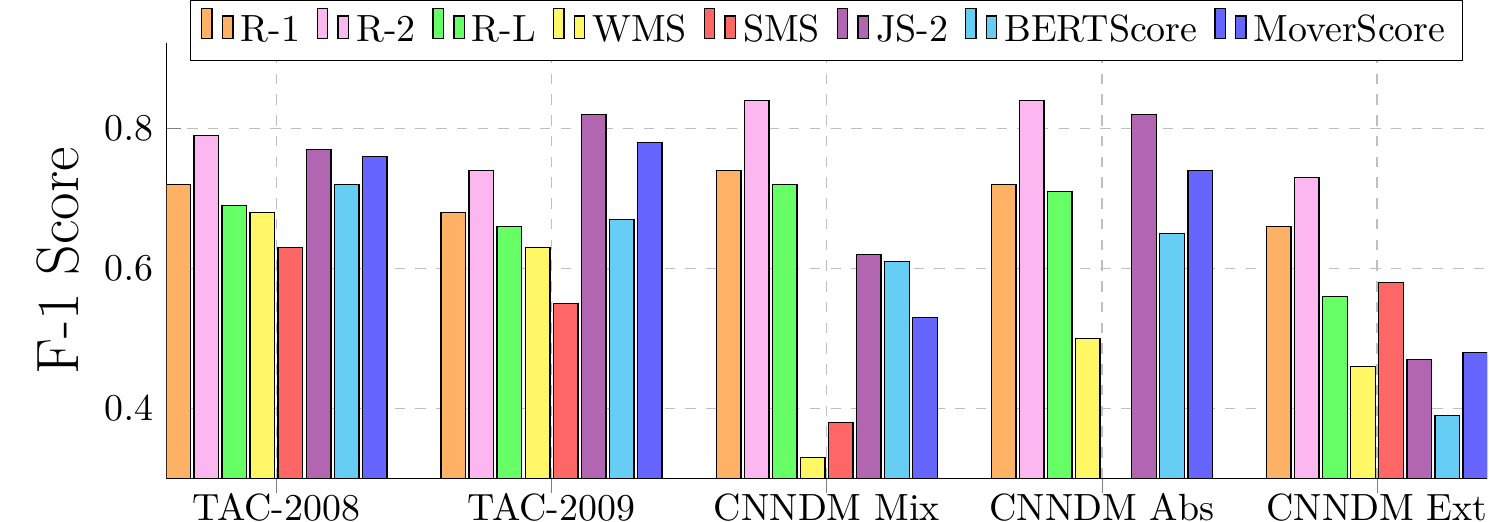}
    \caption{\label{fig:bootstrap}F1-Scores with bootstrapping (Sec.~\ref{sec:bootstrap}).}
\end{figure}

We analyze the reliability of metrics along a useful dimension: \textit{can metrics reliably say if one system is significantly better than another}?
Since we only have 100 annotated summaries to compare any two systems, $sys_1$ and $sys_2$, we use paired bootstrap resampling, to test with statistical significance if $sys_1$ is better than $sys_2$ according to metric $m$~\cite{bootstrap_paper,dror-etal-2018-hitchhikers}.
We take all $\binom{J}{2}$ pairs of systems and compare their mean human score (Eqn.~\ref{eqn:human_score}) using paired bootstrap resampling. We assign a label $y_{true} = 1$ if $sys_1$ is better than $sys_2$ with 95\% confidence, $y_{true} = 2$ for vice-versa and $y_{true} = 0$ if the confidence is below 95\%. We treat this as the ground truth label of the pair $(sys_1, sys_2)$. This process is then repeated for all metrics, to get a ``prediction", $y_{pred}^m$ from each metric $m$ for the same $\binom{J}{2}$ pairs. If $m$ is a good proxy for human judgments, the F1 score~\cite{goutte2005probabilistic} between $y_{pred}^m$ and $y_{true}$ should be high. We calculate the weighted macro F1 score for all metrics and view them in Fig. \ref{fig:bootstrap}.

We find that ROUGE based metrics perform moderately well in this task. R-2 performs the best on \texttt{CNNDM} datasets. While on the \texttt{TAC 2009} dataset, JS-2 achieves the highest F1 score, its performance is low on \texttt{CNNDM Ext}.

\noindent
\textbf{Takeaway:} Different metrics are better suited for different datasets. For example, on the \texttt{CNNDM} datasets, we recommend using R-2 while, on the TAC datasets, we recommend using JS-2. 

\subsection{Exp-IV: Evaluating Summaries}
\label{sec:summ_level_human}

In addition to comparing systems, real-world application scenarios also require metrics to reliably compare multiple summaries of a document. For example, top-scoring reinforcement learning based summarization systems~\cite{bohm2019better} and the current state-of-the-art extractive system~\cite{zhong2020extractive} heavily rely on summary-level reward scores to guide the optimization process.

In this experiment, we ask the question:
\textit{how well do different metrics perform at the summary level, i.e. in comparing system summaries generated from the same document?}
We use Eq.~\ref{eqn:summ_level_corr} to calculate \textit{Pearson} correlation between different metrics and human judgments for different datasets and collected system outputs.
 \begin{figure}
    \centering
    \subfloat[Summary-level \textit{Pearson} correlation with human scores.]{
    \includegraphics[width=0.95\linewidth]{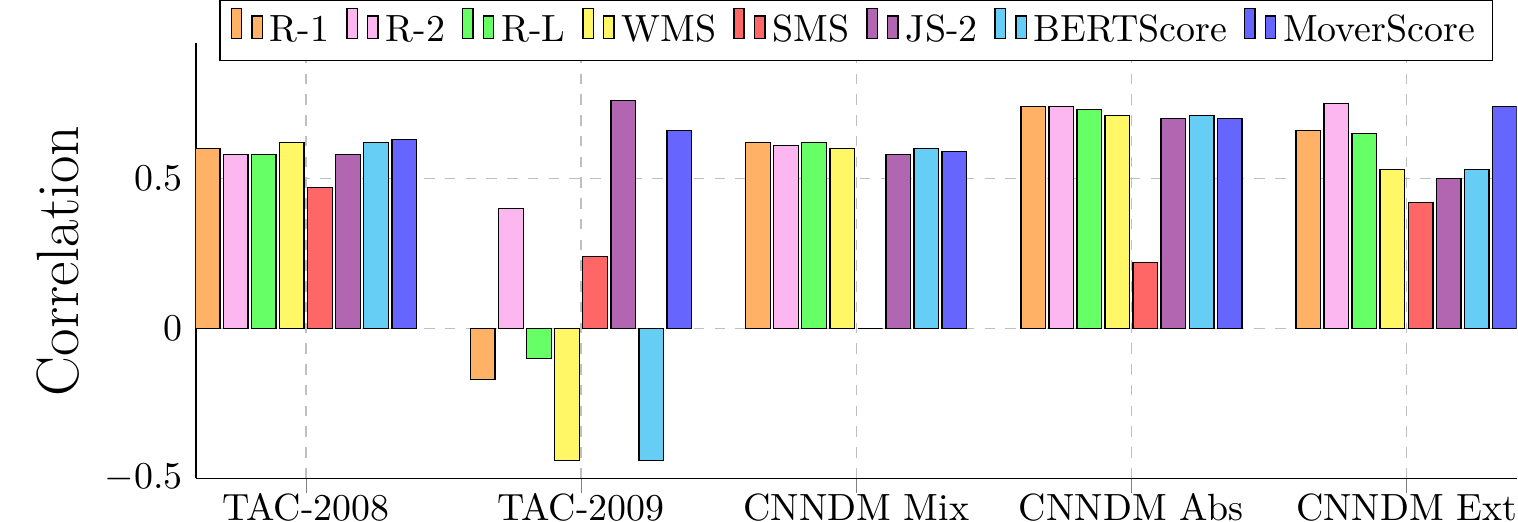}
    \label{fig:summ_level_human}
    }
    
    \subfloat[Difference between system-level and summary-level \textit{Pearson} correlation.]{
    \includegraphics[width=0.95\linewidth]{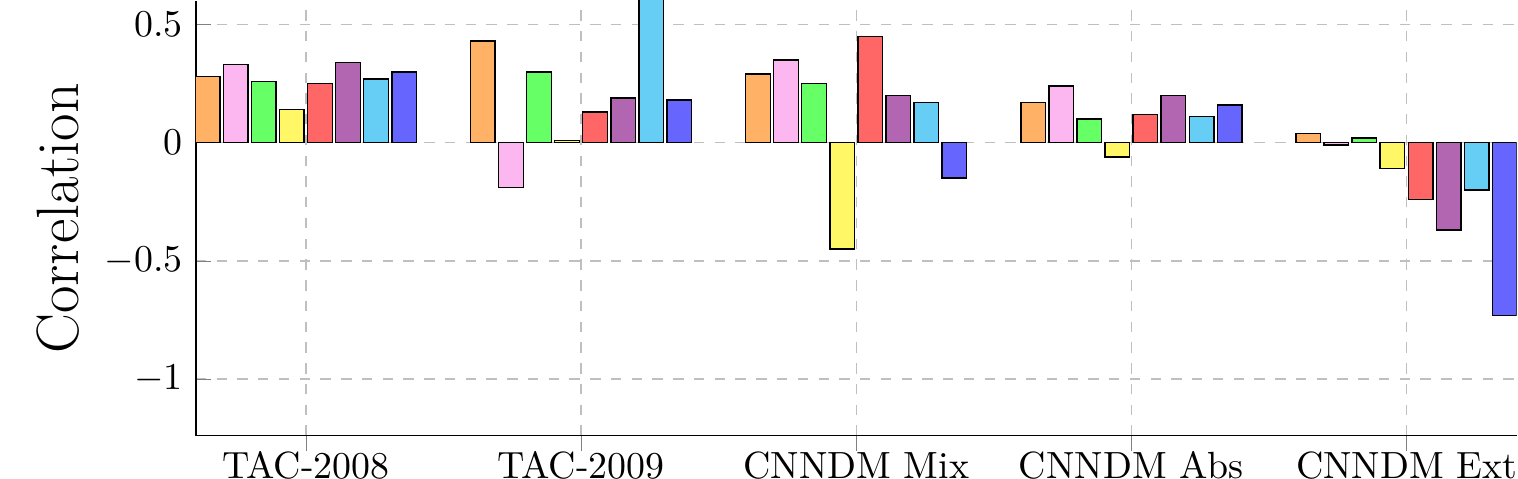}
    \label{fig:sys_summ_level}
    }
    \caption{\label{fig:summ_level_all}\textit{Pearson} correlation between metrics and human judgments across different datasets (Sec.~\ref{sec:summ_level_human}).}
\end{figure}
Our observations are summarized in Fig.~\ref{fig:summ_level_all}. We find that:

\noindent (1) As compared to semantic matching metrics, R-1, R-2 and R-L have lower correlations on the \texttt{TAC} datasets but are strong indicators of good summaries especially for extractive summaries on the \texttt{CNNDM} dataset.

\noindent (2) Notably, BERTScore, WMS, R-1 and R-L have \textit{negative} correlations on \texttt{TAC-2009} but perform moderately well on other datasets including \texttt{CNNDM}.

\noindent
(3) Previous meta-evaluation studies~\cite{novikova-etal-2017-need,peyrard_s3,chaganty2018price} conclude that automatic metrics tend to correlate well with humans at the system level but have poor correlations at the instance (here summary) level. We find this observation only holds on \texttt{TAC-2008}. Some metrics' summary-level correlations can outperform system-level on the \texttt{CNNDM} dataset as shown in Fig.~\ref{fig:sys_summ_level} (bins below $y=0$). Notably, MoverScore has a correlation of only 0.05 on \texttt{CNNDM Ext} at the system level but 0.74 at the summary level.

\noindent \textbf{Takeaway:} Meta-evaluations of metrics on the old \texttt{TAC} datasets show significantly different trends than meta-evaluation on modern systems and datasets. 
Even though some metrics might be good at comparing summaries, they may point in the wrong direction when comparing systems.
Moreover, some metrics show poor generalization ability to different datasets (e.g. BERTScore on \texttt{TAC-2009} vs other datasets). This highlights the need for empirically testing the efficacy of different automatic metrics in evaluating summaries on multiple datasets.

\section{Related Work}

This work is connected to the following threads of topics in text summarization.


\noindent
\textbf{Human Judgment Collection}
Despite many approaches to the acquisition of human judgment \cite{chaganty2018price,nenkova-passonneau-2004-pyramid-og,litepyramids-shapira-etal-2019-crowdsourcing,fan-etal-2018-controllable}, \textit{Pyramid} \cite{nenkova-passonneau-2004-pyramid-og} has been a mainstream method to meta-evaluate various automatic metrics.
Specifically, Pyramid provides a robust technique for evaluating content selection by exhaustively obtaining a set of Semantic Content Units (SCUs) from a set of references, and then scoring system summaries on how many SCUs can be inferred from them.
Recently, \citet{litepyramids-shapira-etal-2019-crowdsourcing} proposed a lightweight and crowdsourceable version of the original Pyramid, and demonstrated it on the DUC 2005 \cite{Dang05overviewof} and 2006 \cite{Dang06overviewof} multi-document summarization datasets.
In this paper, our human evaluation methodology is based on the Pyramid \cite{nenkova-passonneau-2004-pyramid-og} and LitePyramids \cite{litepyramids-shapira-etal-2019-crowdsourcing} techniques.
\citet{chaganty2018price} also obtain human evaluations on system summaries on the CNNDM dataset, but with a focus on language quality of summaries.
In comparison, our work is focused on evaluating content selection. Our work also covers more systems than their study (11 extractive + 14 abstractive vs. 4 abstractive). \\

\noindent \textbf{Meta-evaluation with Human Judgment}
The effectiveness of different automatic metrics - ROUGE-2~\cite{lin2004rouge}, ROUGE-L~\cite{lin2004rouge}, ROUGE-WE~\cite{ng-abrecht-2015-better}, JS-2~\cite{js2} and S3~\cite{peyrard_s3}
is commonly evaluated based on their correlation with human judgments (e.g., on the TAC-2008~\cite{tac2008} and TAC-2009~\cite{tac2009} datasets).
As an important supplementary technique to meta-evaluation, \citet{graham-2015-evaluating} advocate for the use of a significance test, William's test~\cite{williams-test}, to measure the improved correlations of a metric with human scores and show that the popular variant of ROUGE (mean ROUGE-2 score) is sub-optimal. 
Unlike these works, instead of proposing a new metric, in this paper, we upgrade the meta-evaluation environment by introducing a sizeable human judgment dataset evaluating current top-scoring systems and mainstream datasets. And then, we re-evaluate diverse metrics at both system-level and summary-level settings.
\citep{novikova-etal-2017-need} also analyzes existing metrics, but they only focus on dialog generation.

\section{Implications and Future Directions}

Our work not only diagnoses the limitations of current metrics but also highlights the importance of upgrading the existing meta-evaluation testbed, keeping it up-to-date with the rapid development of systems and datasets. In closing, we highlight some potential future directions:
(1) The choice of metrics depends not only on different tasks (e.g, summarization, translation) but also on different datasets (e.g., \texttt{TAC}, \texttt{CNNDM}) and application scenarios (e.g, system-level, summary-level).
Future works on meta-evaluation should investigate the effect of these settings on the performance of metrics.
(2) Metrics easily overfit on limited datasets. Multi-dataset meta-evaluation can help us better understand each metric's peculiarity, therefore achieving a better choice of metrics under diverse scenarios.
(3) Our collected human judgments can be used as supervision to instantiate the most recently-proposed \textit{pretrain-then-finetune} framework (originally for machine translation) \cite{sellam2020bleurt}, learning a robust metric for text summarization.


\section*{Acknowledgements}
We sincerely thank all authors of the systems that we used in this work for sharing their systems' outputs.






\bibliography{emnlp2020}

\begin{thebibliography}{55}
\expandafter\ifx\csname natexlab\endcsname\relax\def\natexlab#1{#1}\fi

\bibitem[{Bojar et~al.(2016)Bojar, Graham, Kamran, and
  Stanojevi{\'c}}]{bojar-etal-2016-wmt-results}
Ond{\v{r}}ej Bojar, Yvette Graham, Amir Kamran, and Milo{\v{s}} Stanojevi{\'c}.
  2016.
\newblock \href {https://doi.org/10.18653/v1/W16-2302} {Results of the {WMT}16
  metrics shared task}.
\newblock In \emph{Proceedings of the First Conference on Machine Translation:
  Volume 2, Shared Task Papers}, pages 199--231, Berlin, Germany. Association
  for Computational Linguistics.

\bibitem[{Böhm et~al.(2019)Böhm, Gao, Meyer, Shapira, Dagan, and
  Gurevych}]{bohm2019better}
Florian Böhm, Yang Gao, Christian~M. Meyer, Ori Shapira, Ido Dagan, and Iryna
  Gurevych. 2019.
\newblock \href {http://arxiv.org/abs/1909.01214} {Better rewards yield better
  summaries: Learning to summarise without references}.

\bibitem[{Chaganty et~al.(2018)Chaganty, Mussman, and
  Liang}]{chaganty2018price}
Arun~Tejasvi Chaganty, Stephen Mussman, and Percy Liang. 2018.
\newblock \href {http://arxiv.org/abs/1807.02202} {The price of debiasing
  automatic metrics in natural language evaluation}.

\bibitem[{Chen and Bansal(2018)}]{chen-bansal-2018-fast-abs}
Yen-Chun Chen and Mohit Bansal. 2018.
\newblock \href {https://doi.org/10.18653/v1/P18-1063} {Fast abstractive
  summarization with reinforce-selected sentence rewriting}.
\newblock In \emph{Proceedings of the 56th Annual Meeting of the Association
  for Computational Linguistics (Volume 1: Long Papers)}, pages 675--686,
  Melbourne, Australia. Association for Computational Linguistics.

\bibitem[{Clark et~al.(2019)Clark, Celikyilmaz, and Smith}]{clark2019sentence}
Elizabeth Clark, Asli Celikyilmaz, and Noah~A Smith. 2019.
\newblock Sentence mover’s similarity: Automatic evaluation for
  multi-sentence texts.
\newblock In \emph{Proceedings of the 57th Annual Meeting of the Association
  for Computational Linguistics}, pages 2748--2760.

\bibitem[{Dang and Owczarzak(2008)}]{tac2008}
Hoa Dang and Karolina Owczarzak. 2008.
\newblock Overview of the tac 2008 update summarization task.
\newblock In \emph{Proceedings of the First Text Analysis Conference (TAC
  2008)}, pages 1 -- 16.

\bibitem[{Dang and Owczarzak(2009)}]{tac2009}
Hoa Dang and Karolina Owczarzak. 2009.
\newblock Overview of the tac 2009 summarization track.
\newblock In \emph{Proceedings of the First Text Analysis Conference (TAC
  2009)}, pages 1 -- 16.

\bibitem[{Dang(2005)}]{Dang05overviewof}
Hoa~Trang Dang. 2005.
\newblock Overview of duc 2005.
\newblock In \emph{In Proceedings of the Document Understanding Conf. Wksp.
  2005 (DUC 2005) at the Human Language Technology Conf./Conf. on Empirical
  Methods in Natural Language Processing (HLT/EMNLP}.

\bibitem[{Dang(2006)}]{Dang06overviewof}
Hoa~Trang Dang. 2006.
\newblock Overview of duc 2006.
\newblock In \emph{In Proceedings of HLT-NAACL 2006}.

\bibitem[{Dong et~al.(2019)Dong, Yang, Wang, Wei, Liu, Wang, Gao, Zhou, and
  Hon}]{dong2019unified}
Li~Dong, Nan Yang, Wenhui Wang, Furu Wei, Xiaodong Liu, Yu~Wang, Jianfeng Gao,
  Ming Zhou, and Hsiao-Wuen Hon. 2019.
\newblock Unified language model pre-training for natural language
  understanding and generation.
\newblock In \emph{Advances in Neural Information Processing Systems}, pages
  13042--13054.

\bibitem[{Dong et~al.(2018)Dong, Shen, Crawford, van Hoof, and
  Cheung}]{dong-etal-2018-banditsum}
Yue Dong, Yikang Shen, Eric Crawford, Herke van Hoof, and Jackie Chi~Kit
  Cheung. 2018.
\newblock \href {https://doi.org/10.18653/v1/D18-1409} {{B}andit{S}um:
  Extractive summarization as a contextual bandit}.
\newblock In \emph{Proceedings of the 2018 Conference on Empirical Methods in
  Natural Language Processing}, pages 3739--3748, Brussels, Belgium.
  Association for Computational Linguistics.

\bibitem[{Dror et~al.(2018)Dror, Baumer, Shlomov, and
  Reichart}]{dror-etal-2018-hitchhikers}
Rotem Dror, Gili Baumer, Segev Shlomov, and Roi Reichart. 2018.
\newblock \href {https://doi.org/10.18653/v1/P18-1128} {The hitchhiker{'}s
  guide to testing statistical significance in natural language processing}.
\newblock In \emph{Proceedings of the 56th Annual Meeting of the Association
  for Computational Linguistics (Volume 1: Long Papers)}, pages 1383--1392,
  Melbourne, Australia. Association for Computational Linguistics.

\bibitem[{Fan et~al.(2018)Fan, Grangier, and Auli}]{fan-etal-2018-controllable}
Angela Fan, David Grangier, and Michael Auli. 2018.
\newblock \href {https://doi.org/10.18653/v1/W18-2706} {Controllable
  abstractive summarization}.
\newblock In \emph{Proceedings of the 2nd Workshop on Neural Machine
  Translation and Generation}, pages 45--54, Melbourne, Australia. Association
  for Computational Linguistics.

\bibitem[{Gehrmann et~al.(2018)Gehrmann, Deng, and Rush}]{gehrmann2018bottom}
Sebastian Gehrmann, Yuntian Deng, and Alexander Rush. 2018.
\newblock Bottom-up abstractive summarization.
\newblock In \emph{Proceedings of the 2018 Conference on Empirical Methods in
  Natural Language Processing}, pages 4098--4109.

\bibitem[{Goutte and Gaussier(2005)}]{goutte2005probabilistic}
Cyril Goutte and Eric Gaussier. 2005.
\newblock A probabilistic interpretation of precision, recall and f-score, with
  implication for evaluation.
\newblock In \emph{European Conference on Information Retrieval}, pages
  345--359. Springer.

\bibitem[{Graham(2015)}]{graham-2015-evaluating}
Yvette Graham. 2015.
\newblock \href {https://doi.org/10.18653/v1/D15-1013} {Re-evaluating automatic
  summarization with {BLEU} and 192 shades of {ROUGE}}.
\newblock In \emph{Proceedings of the 2015 Conference on Empirical Methods in
  Natural Language Processing}, pages 128--137, Lisbon, Portugal. Association
  for Computational Linguistics.

\bibitem[{Graham and Baldwin(2014)}]{graham-baldwin-2014-testing}
Yvette Graham and Timothy Baldwin. 2014.
\newblock \href {https://doi.org/10.3115/v1/D14-1020} {Testing for significance
  of increased correlation with human judgment}.
\newblock In \emph{Proceedings of the 2014 Conference on Empirical Methods in
  Natural Language Processing ({EMNLP})}, pages 172--176, Doha, Qatar.
  Association for Computational Linguistics.

\bibitem[{Hermann et~al.(2015)Hermann, Kocisky, Grefenstette, Espeholt, Kay,
  Suleyman, and Blunsom}]{hermann2015teaching}
Karl~Moritz Hermann, Tomas Kocisky, Edward Grefenstette, Lasse Espeholt, Will
  Kay, Mustafa Suleyman, and Phil Blunsom. 2015.
\newblock Teaching machines to read and comprehend.
\newblock In \emph{Advances in Neural Information Processing Systems}, pages
  1684--1692.

\bibitem[{Kedzie et~al.(2018)Kedzie, McKeown, and
  Daume~III}]{kedzie2018content}
Chris Kedzie, Kathleen McKeown, and Hal Daume~III. 2018.
\newblock Content selection in deep learning models of summarization.
\newblock In \emph{Proceedings of the 2018 Conference on Empirical Methods in
  Natural Language Processing}, pages 1818--1828.

\bibitem[{Koehn(2004)}]{bootstrap_paper}
Philipp Koehn. 2004.
\newblock \href {https://www.aclweb.org/anthology/W04-3250} {Statistical
  significance tests for machine translation evaluation}.
\newblock In \emph{Proceedings of the 2004 Conference on Empirical Methods in
  Natural Language Processing}, pages 388--395, Barcelona, Spain. Association
  for Computational Linguistics.

\bibitem[{Krippendorff(2011)}]{krippendorff2011computing}
Klaus Krippendorff. 2011.
\newblock Computing krippendorff's alpha-reliability.

\bibitem[{Kusner et~al.(2015)Kusner, Sun, Kolkin, and
  Weinberger}]{kusner2015word}
Matt Kusner, Yu~Sun, Nicholas Kolkin, and Kilian Weinberger. 2015.
\newblock From word embeddings to document distances.
\newblock In \emph{International conference on machine learning}, pages
  957--966.

\bibitem[{Lee~Rodgers(1988)}]{lee1988thirteen}
W~Alan Lee~Rodgers. 1988.
\newblock Thirteen ways to look at the correlation coefficient.
\newblock \emph{The American Statistician}, 42(1):59--66.

\bibitem[{Lewis et~al.(2019)Lewis, Liu, Goyal, Ghazvininejad, Mohamed, Levy,
  Stoyanov, and Zettlemoyer}]{lewis2019bart}
Mike Lewis, Yinhan Liu, Naman Goyal, Marjan Ghazvininejad, Abdelrahman Mohamed,
  Omer Levy, Ves Stoyanov, and Luke Zettlemoyer. 2019.
\newblock Bart: Denoising sequence-to-sequence pre-training for natural
  language generation, translation, and comprehension.
\newblock \emph{arXiv preprint arXiv:1910.13461}.

\bibitem[{Lin(2004)}]{lin2004rouge}
Chin-Yew Lin. 2004.
\newblock Rouge: A package for automatic evaluation of summaries.
\newblock \emph{Text Summarization Branches Out}.

\bibitem[{Lin et~al.(2006)Lin, Cao, Gao, and Nie}]{lin-etal-2006-information}
Chin-Yew Lin, Guihong Cao, Jianfeng Gao, and Jian-Yun Nie. 2006.
\newblock \href {https://www.aclweb.org/anthology/N06-1059} {An
  information-theoretic approach to automatic evaluation of summaries}.
\newblock In \emph{Proceedings of the Human Language Technology Conference of
  the {NAACL}, Main Conference}, pages 463--470, New York City, USA.
  Association for Computational Linguistics.

\bibitem[{Lin and Och(2004)}]{lin-och-2004-orange}
Chin-Yew Lin and Franz~Josef Och. 2004.
\newblock \href {https://www.aclweb.org/anthology/C04-1072} {{ORANGE}: a method
  for evaluating automatic evaluation metrics for machine translation}.
\newblock In \emph{{COLING} 2004: Proceedings of the 20th International
  Conference on Computational Linguistics}, pages 501--507, Geneva,
  Switzerland. COLING.

\bibitem[{Liu and Lapata(2019{\natexlab{a}})}]{liu-lapata-2019-text}
Yang Liu and Mirella Lapata. 2019{\natexlab{a}}.
\newblock \href {https://doi.org/10.18653/v1/D19-1387} {Text summarization with
  pretrained encoders}.
\newblock In \emph{Proceedings of the 2019 Conference on Empirical Methods in
  Natural Language Processing and the 9th International Joint Conference on
  Natural Language Processing (EMNLP-IJCNLP)}, pages 3730--3740, Hong Kong,
  China. Association for Computational Linguistics.

\bibitem[{Liu and Lapata(2019{\natexlab{b}})}]{liu2019text-presumm}
Yang Liu and Mirella Lapata. 2019{\natexlab{b}}.
\newblock \href {http://arxiv.org/abs/1908.08345} {Text summarization with
  pretrained encoders}.

\bibitem[{Louis and Nenkova(2013)}]{js2}
Annie Louis and Ani Nenkova. 2013.
\newblock \href {https://doi.org/10.1162/COLI_a_00123} {Automatically assessing
  machine summary content without a gold standard}.
\newblock \emph{Computational Linguistics}, 39(2):267--300.

\bibitem[{Manning et~al.(2014)Manning, Surdeanu, Bauer, Finkel, Bethard, and
  McClosky}]{manning-EtAl-2014-CoreNLP}
Christopher~D. Manning, Mihai Surdeanu, John Bauer, Jenny Finkel, Steven~J.
  Bethard, and David McClosky. 2014.
\newblock \href {http://www.aclweb.org/anthology/P/P14/P14-5010} {The
  {Stanford} {CoreNLP} natural language processing toolkit}.
\newblock In \emph{Association for Computational Linguistics (ACL) System
  Demonstrations}, pages 55--60.

\bibitem[{Nallapati et~al.(2016)Nallapati, Zhou, dos Santos, glar
  Gul{\c{c}}ehre, and Xiang}]{nallapati2016abstractive}
Ramesh Nallapati, Bowen Zhou, Cicero dos Santos, {\c{C}}a~glar Gul{\c{c}}ehre,
  and Bing Xiang. 2016.
\newblock Abstractive text summarization using sequence-to-sequence rnns and
  beyond.
\newblock \emph{CoNLL 2016}, page 280.

\bibitem[{Narayan et~al.(2018)Narayan, Cohen, and
  Lapata}]{narayan-etal-2018-ranking}
Shashi Narayan, Shay~B. Cohen, and Mirella Lapata. 2018.
\newblock \href {https://doi.org/10.18653/v1/N18-1158} {Ranking sentences for
  extractive summarization with reinforcement learning}.
\newblock In \emph{Proceedings of the 2018 Conference of the North {A}merican
  Chapter of the Association for Computational Linguistics: Human Language
  Technologies, Volume 1 (Long Papers)}, pages 1747--1759, New Orleans,
  Louisiana. Association for Computational Linguistics.

\bibitem[{Nenkova and Passonneau(2004)}]{nenkova-passonneau-2004-pyramid-og}
Ani Nenkova and Rebecca Passonneau. 2004.
\newblock \href {https://www.aclweb.org/anthology/N04-1019} {Evaluating content
  selection in summarization: The pyramid method}.
\newblock In \emph{Proceedings of the Human Language Technology Conference of
  the North {A}merican Chapter of the Association for Computational
  Linguistics: {HLT}-{NAACL} 2004}, pages 145--152, Boston, Massachusetts, USA.
  Association for Computational Linguistics.

\bibitem[{Ng and Abrecht(2015)}]{ng-abrecht-2015-better}
Jun-Ping Ng and Viktoria Abrecht. 2015.
\newblock \href {https://doi.org/10.18653/v1/D15-1222} {Better summarization
  evaluation with word embeddings for {ROUGE}}.
\newblock In \emph{Proceedings of the 2015 Conference on Empirical Methods in
  Natural Language Processing}, pages 1925--1930, Lisbon, Portugal. Association
  for Computational Linguistics.

\bibitem[{Novikova et~al.(2017)Novikova, Du{\v{s}}ek, Cercas~Curry, and
  Rieser}]{novikova-etal-2017-need}
Jekaterina Novikova, Ond{\v{r}}ej Du{\v{s}}ek, Amanda Cercas~Curry, and Verena
  Rieser. 2017.
\newblock \href {https://doi.org/10.18653/v1/D17-1238} {Why we need new
  evaluation metrics for {NLG}}.
\newblock In \emph{Proceedings of the 2017 Conference on Empirical Methods in
  Natural Language Processing}, pages 2241--2252, Copenhagen, Denmark.
  Association for Computational Linguistics.

\bibitem[{Peyrard(2019)}]{peyrard-2019-studying}
Maxime Peyrard. 2019.
\newblock \href {https://doi.org/10.18653/v1/P19-1502} {Studying summarization
  evaluation metrics in the appropriate scoring range}.
\newblock In \emph{Proceedings of the 57th Annual Meeting of the Association
  for Computational Linguistics}, pages 5093--5100, Florence, Italy.
  Association for Computational Linguistics.

\bibitem[{Peyrard et~al.(2017)Peyrard, Botschen, and Gurevych}]{peyrard_s3}
Maxime Peyrard, Teresa Botschen, and Iryna Gurevych. 2017.
\newblock \href {https://doi.org/10.18653/v1/W17-4510} {Learning to score
  system summaries for better content selection evaluation.}
\newblock In \emph{Proceedings of the Workshop on New Frontiers in
  Summarization}, pages 74--84, Copenhagen, Denmark. Association for
  Computational Linguistics.

\bibitem[{Raffel et~al.(2019)Raffel, Shazeer, Roberts, Lee, Narang, Matena,
  Zhou, Li, and Liu}]{raffel2019exploring-t5}
Colin Raffel, Noam Shazeer, Adam Roberts, Katherine Lee, Sharan Narang, Michael
  Matena, Yanqi Zhou, Wei Li, and Peter~J. Liu. 2019.
\newblock \href {http://arxiv.org/abs/1910.10683} {Exploring the limits of
  transfer learning with a unified text-to-text transformer}.

\bibitem[{Rankel et~al.(2013)Rankel, Conroy, Dang, and
  Nenkova}]{rankel-etal-2013-decade}
Peter~A. Rankel, John~M. Conroy, Hoa~Trang Dang, and Ani Nenkova. 2013.
\newblock \href {https://www.aclweb.org/anthology/P13-2024} {A decade of
  automatic content evaluation of news summaries: Reassessing the state of the
  art}.
\newblock In \emph{Proceedings of the 51st Annual Meeting of the Association
  for Computational Linguistics (Volume 2: Short Papers)}, pages 131--136,
  Sofia, Bulgaria. Association for Computational Linguistics.

\bibitem[{See et~al.(2017)See, Liu, and Manning}]{see-etal-2017-get}
Abigail See, Peter~J. Liu, and Christopher~D. Manning. 2017.
\newblock \href {https://doi.org/10.18653/v1/P17-1099} {Get to the point:
  Summarization with pointer-generator networks}.
\newblock In \emph{Proceedings of the 55th Annual Meeting of the Association
  for Computational Linguistics (Volume 1: Long Papers)}, pages 1073--1083,
  Vancouver, Canada. Association for Computational Linguistics.

\bibitem[{Sellam et~al.(2020)Sellam, Das, and Parikh}]{sellam2020bleurt}
Thibault Sellam, Dipanjan Das, and Ankur~P Parikh. 2020.
\newblock Bleurt: Learning robust metrics for text generation.
\newblock \emph{arXiv preprint arXiv:2004.04696}.

\bibitem[{Shapira et~al.(2019)Shapira, Gabay, Gao, Ronen, Pasunuru, Bansal,
  Amsterdamer, and Dagan}]{litepyramids-shapira-etal-2019-crowdsourcing}
Ori Shapira, David Gabay, Yang Gao, Hadar Ronen, Ramakanth Pasunuru, Mohit
  Bansal, Yael Amsterdamer, and Ido Dagan. 2019.
\newblock \href {https://doi.org/10.18653/v1/N19-1072} {Crowdsourcing
  lightweight pyramids for manual summary evaluation}.
\newblock In \emph{Proceedings of the 2019 Conference of the North {A}merican
  Chapter of the Association for Computational Linguistics: Human Language
  Technologies, Volume 1 (Long and Short Papers)}, pages 682--687, Minneapolis,
  Minnesota. Association for Computational Linguistics.

\bibitem[{{Virtanen} et~al.(2020){Virtanen}, {Gommers}, {Oliphant},
  {Haberland}, {Reddy}, {Cournapeau}, {Burovski}, {Peterson}, {Weckesser},
  {Bright}, {van der Walt}, {Brett}, {Wilson}, {Jarrod Millman}, {Mayorov},
  {Nelson}, {Jones}, {Kern}, {Larson}, {Carey}, {Polat}, {Feng}, {Moore}, {Vand
  erPlas}, {Laxalde}, {Perktold}, {Cimrman}, {Henriksen}, {Quintero}, {Harris},
  {Archibald}, {Ribeiro}, {Pedregosa}, {van Mulbregt}, and
  {Contributors}}]{scipy-2020}
Pauli {Virtanen}, Ralf {Gommers}, Travis~E. {Oliphant}, Matt {Haberland}, Tyler
  {Reddy}, David {Cournapeau}, Evgeni {Burovski}, Pearu {Peterson}, Warren
  {Weckesser}, Jonathan {Bright}, St{\'e}fan~J. {van der Walt}, Matthew
  {Brett}, Joshua {Wilson}, K.~{Jarrod Millman}, Nikolay {Mayorov}, Andrew
  R.~J. {Nelson}, Eric {Jones}, Robert {Kern}, Eric {Larson}, CJ~{Carey},
  {\.I}lhan {Polat}, Yu~{Feng}, Eric~W. {Moore}, Jake {Vand erPlas}, Denis
  {Laxalde}, Josef {Perktold}, Robert {Cimrman}, Ian {Henriksen}, E.~A.
  {Quintero}, Charles~R {Harris}, Anne~M. {Archibald}, Ant{\^o}nio~H.
  {Ribeiro}, Fabian {Pedregosa}, Paul {van Mulbregt}, and SciPy 1.~0
  {Contributors}. 2020.
\newblock \href {https://doi.org/https://doi.org/10.1038/s41592-019-0686-2}
  {{SciPy 1.0: Fundamental Algorithms for Scientific Computing in Python}}.
\newblock \emph{Nature Methods}, 17:261--272.

\bibitem[{Wang et~al.(2020)Wang, Liu, Zheng, Qiu, and
  Huang}]{wang2020heterogeneous}
Danqing Wang, Pengfei Liu, Yining Zheng, Xipeng Qiu, and Xuanjing Huang. 2020.
\newblock Heterogeneous graph neural networks for extractive document
  summarization.
\newblock \emph{arXiv preprint arXiv:2004.12393}.

\bibitem[{Williams(1959)}]{williams-test}
Evan~J. Williams. 1959.
\newblock Regression analysis.
\newblock \emph{Wiley, New York}, 14.

\bibitem[{Yoon et~al.(2020)Yoon, Yeo, Jeong, Yi, and Kang}]{yoon2020learning}
Wonjin Yoon, Yoon~Sun Yeo, Minbyul Jeong, Bong-Jun Yi, and Jaewoo Kang. 2020.
\newblock Learning by semantic similarity makes abstractive summarization
  better.
\newblock \emph{arXiv preprint arXiv:2002.07767}.

\bibitem[{Zhang et~al.(2019{\natexlab{a}})Zhang, Gong, Yan, Duan, Xu, Wang,
  Gong, and Zhou}]{zhang2019pretraining}
Haoyu Zhang, Yeyun Gong, Yu~Yan, Nan Duan, Jianjun Xu, Ji~Wang, Ming Gong, and
  Ming Zhou. 2019{\natexlab{a}}.
\newblock Pretraining-based natural language generation for text summarization.
\newblock \emph{arXiv preprint arXiv:1902.09243}.

\bibitem[{Zhang et~al.(2020)Zhang, Kishore, Wu, Weinberger, and
  Artzi}]{bert-score}
Tianyi Zhang, Varsha Kishore, Felix Wu, Kilian~Q. Weinberger, and Yoav Artzi.
  2020.
\newblock \href {https://openreview.net/forum?id=SkeHuCVFDr} {Bertscore:
  Evaluating text generation with bert}.
\newblock In \emph{International Conference on Learning Representations}.

\bibitem[{Zhang et~al.(2018)Zhang, Lapata, Wei, and
  Zhou}]{zhang-etal-2018-neural-latent}
Xingxing Zhang, Mirella Lapata, Furu Wei, and Ming Zhou. 2018.
\newblock \href {https://doi.org/10.18653/v1/D18-1088} {Neural latent
  extractive document summarization}.
\newblock In \emph{Proceedings of the 2018 Conference on Empirical Methods in
  Natural Language Processing}, pages 779--784, Brussels, Belgium. Association
  for Computational Linguistics.

\bibitem[{Zhang et~al.(2019{\natexlab{b}})Zhang, Wei, and
  Zhou}]{zhang-etal-2019-hibert}
Xingxing Zhang, Furu Wei, and Ming Zhou. 2019{\natexlab{b}}.
\newblock \href {https://doi.org/10.18653/v1/P19-1499} {{HIBERT}: Document
  level pre-training of hierarchical bidirectional transformers for document
  summarization}.
\newblock In \emph{Proceedings of the 57th Annual Meeting of the Association
  for Computational Linguistics}, pages 5059--5069, Florence, Italy.
  Association for Computational Linguistics.

\bibitem[{Zhao et~al.(2019)Zhao, Peyrard, Liu, Gao, Meyer, and
  Eger}]{zhao-etal-2019-moverscore}
Wei Zhao, Maxime Peyrard, Fei Liu, Yang Gao, Christian~M. Meyer, and Steffen
  Eger. 2019.
\newblock \href {https://doi.org/10.18653/v1/D19-1053} {{M}over{S}core: Text
  generation evaluating with contextualized embeddings and earth mover
  distance}.
\newblock In \emph{Proceedings of the 2019 Conference on Empirical Methods in
  Natural Language Processing and the 9th International Joint Conference on
  Natural Language Processing (EMNLP-IJCNLP)}, pages 563--578, Hong Kong,
  China. Association for Computational Linguistics.

\bibitem[{Zhong et~al.(2020)Zhong, Liu, Chen, Wang, Qiu, and
  Huang}]{zhong2020extractive}
Ming Zhong, Pengfei Liu, Yiran Chen, Danqing Wang, Xipeng Qiu, and Xuanjing
  Huang. 2020.
\newblock Extractive summarization as text matching.
\newblock \emph{arXiv preprint arXiv:2004.08795}.

\bibitem[{Zhong et~al.(2019)Zhong, Liu, Wang, Qiu, and
  Huang}]{zhong2019searching}
Ming Zhong, Pengfei Liu, Danqing Wang, Xipeng Qiu, and Xuan-Jing Huang. 2019.
\newblock Searching for effective neural extractive summarization: What works
  and what’s next.
\newblock In \emph{Proceedings of the 57th Conference of the Association for
  Computational Linguistics}, pages 1049--1058.

\bibitem[{Zhou et~al.(2018)Zhou, Yang, Wei, Huang, Zhou, and
  Zhao}]{zhou-etal-2018-neural}
Qingyu Zhou, Nan Yang, Furu Wei, Shaohan Huang, Ming Zhou, and Tiejun Zhao.
  2018.
\newblock \href {https://doi.org/10.18653/v1/P18-1061} {Neural document
  summarization by jointly learning to score and select sentences}.
\newblock In \emph{Proceedings of the 56th Annual Meeting of the Association
  for Computational Linguistics (Volume 1: Long Papers)}, pages 654--663,
  Melbourne, Australia. Association for Computational Linguistics.

\end{thebibliography}
\bibliographystyle{acl_natbib}

\newpage
\appendix
\section{Appendices}

\label{sec:appendix}

\subsection{Sampling Methodology}
\label{sec:sampling_algo}
Please see Algorithm~\ref{alg:sampling}.

\begin{algorithm}[ht!]

\caption{\label{alg:sampling}Sampling Methodology}
\SetAlgoLined
\KwData{$(d_i, r_i, S_i) \in D$ where $D$ is \texttt{CNNDM} test set, $d_i$ is source document, $r_i$ is reference summary, and $S_i$ is a set of individual system summaries $s_{ij} \in S_i$. $M$ = [ROUGE-1, ROUGE-2, ROUGE-L, BERTScore, MoverScore]
}
\KwOutput{$D_{out}$: sampled set of documents}
$\mu_{m, i} := mean(\{m(r_i, s_{ij}) \forall s_{ij} \in S_i\})$, $m \in M$ \\
$\sigma_{m, i} := std.dev(\{m(r_i, s_{ij}) \forall s_{ij} \in S_i\})$, $m \in M$ \\
$D_{out} := \{\}$ \\ 
$n_1 := |D| / 5$ \\

\For{$m \in M$}{
$D'$ := $[d_i: d_i \in D]$ sorted by  $\mu_{m,i}$ \\
    \For{$k \in [0, 1, 2, 3, 4]$}{
        $D'_k = D'[i*n_1: (i+1)*n_1]$ \\
        $D''_k = [d_i: d_i \in D'_k]$ sorted by $\sigma_{m,i}$
        $n_2 = |D''_k| / 4$ \\
        \For{$l \in [0, 1, 2, 3]$}{
            $D''_{kl} = D''_k[l*n_2 : (l+1)*n_2]$ \\
            Randomly sample $d_i$ from $D''_{kl}$ \\
            $D_{out} = D_{out} \cup {d_i}$ 
        }
    }
}
\end{algorithm}

\subsection{Exp-I using Kendall's tau correlation}
\begin{figure}[ht!]
    \centering
    \includegraphics[width=\linewidth]{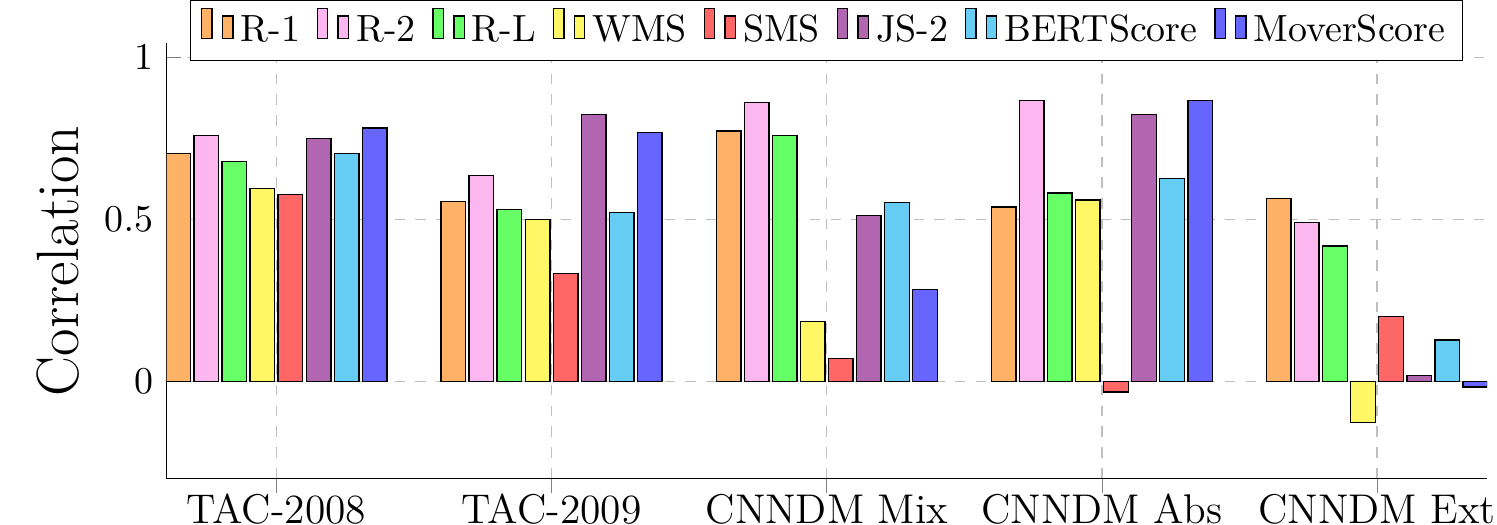}
    \caption{\label{fig:sys_ktau}System-level \textit{Kendall} correlation between metrics and human scores.
    }
\end{figure}

Please see Figure~\ref{fig:sys_ktau} for the system level Kendall's tau correlation between different metrics and human judgements.

\begin{figure}[ht!]
    \centering
    \subfloat[Summary-level \textit{Kendall} correlation with human scores.]{
    \includegraphics[width=0.95\linewidth]{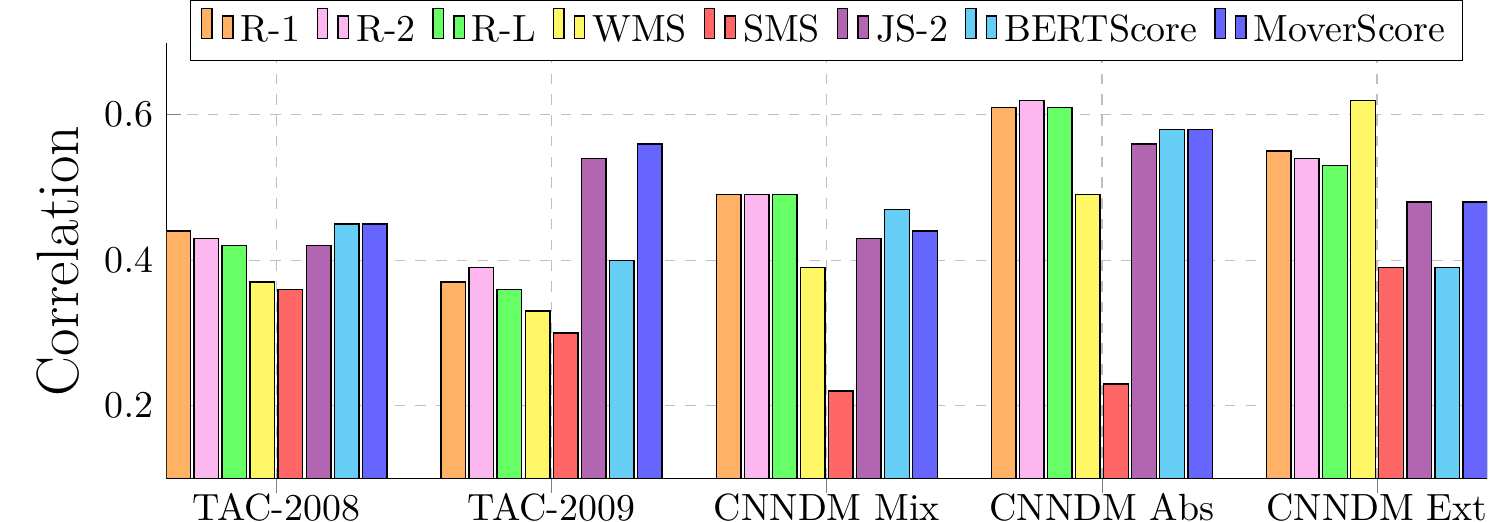}
    \label{fig:summ_level_human_ktau}
    }
    
    \subfloat[Difference between system-level and summary-level \textit{Kendall} correlation.]{
    \includegraphics[width=0.95\linewidth]{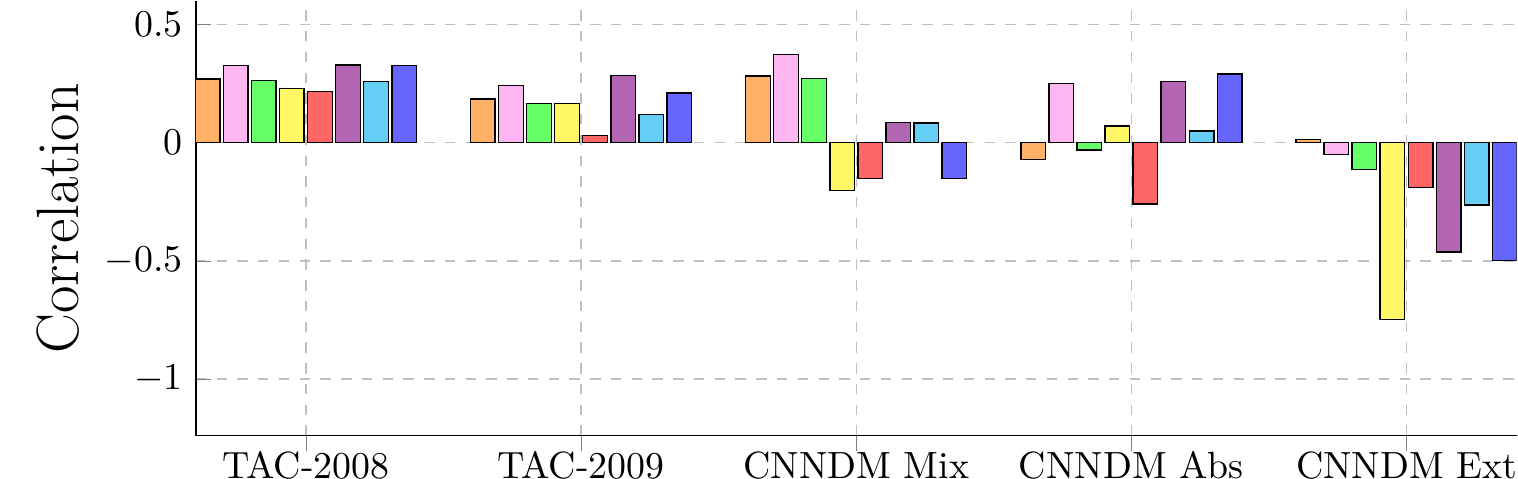}
    \label{fig:sys_summ_level_ktau}
    }
    \caption{\label{fig:summ_level_all_ktau}\textit{Kendall} correlation between metrics and human judgements across different datasets.}
\end{figure}

\subsection{Exp-II using Kendall's tau correlation}
\begin{figure*}[t!]
  \centering
  \subfloat[TAC-2008]{
    \label{fig:tac_8_topk_ktau}
    \includegraphics[width=0.205\linewidth]{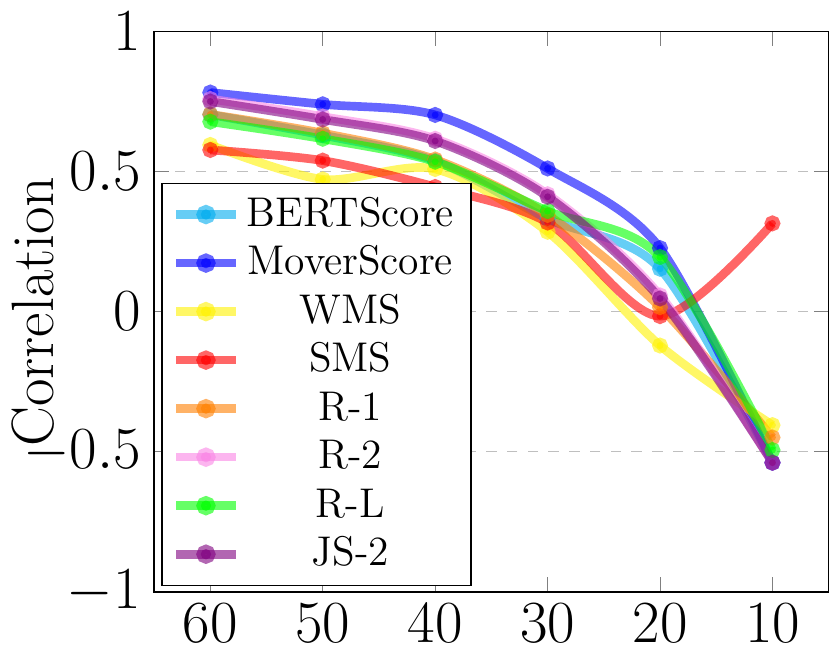}
  }
  \subfloat[TAC-2009]{
    \label{fig:xsum_ktau}
    \includegraphics[width=0.1725\linewidth]{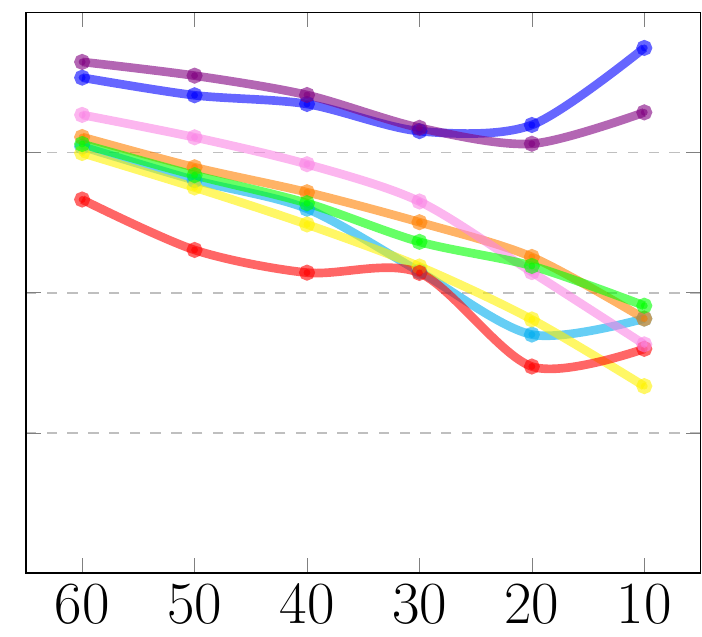}
  }
  \subfloat[CNNDM Mix]{
    \label{fig:cnn_dm_mix_topk_ktau}
    \includegraphics[width=0.1725\linewidth]{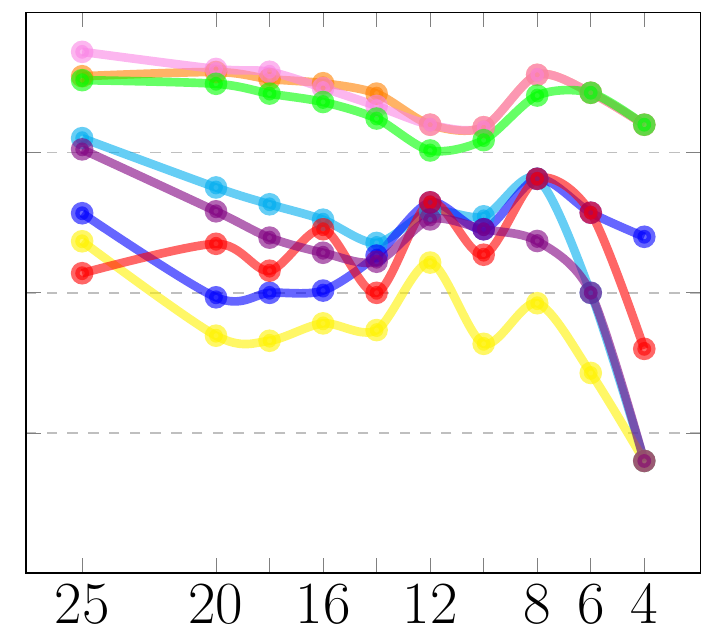}
  }
  \subfloat[CNNDM Abs]{
    \label{fig:cnn_dm_abs_topk_ktau}
    \includegraphics[width=0.1725\linewidth]{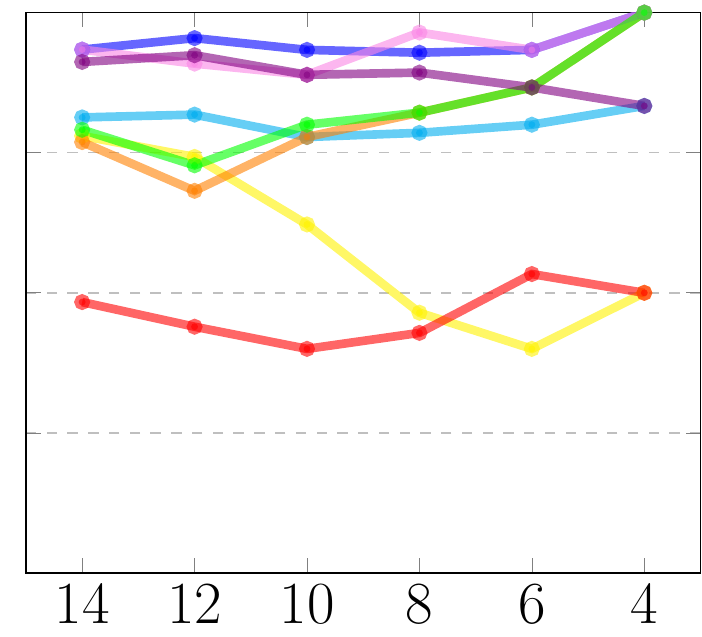}
  }
  \subfloat[CNNDM Ext]{
    \label{fig:cnn_dm_ext_topk_ktau}
    \includegraphics[width=0.1725\linewidth]{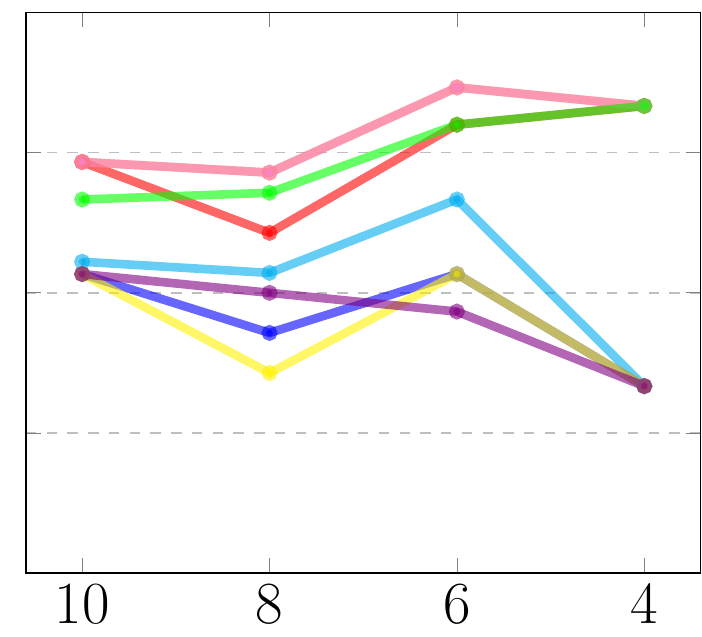}
  }
  
\caption{System-level \textit{Kendall} correlation with humans on top-$k$ systems.
}
 \label{fig:top_k_human_ktau}
\end{figure*}

Please see Figure~\ref{fig:top_k_human_ktau} for the system level Kendall's tau correlation on top-$k$ systems, between different metrics and human judgements.

\subsection{Exp IV using Kendall's tau correlation}
Please see Figure~\ref{fig:summ_level_all_ktau} for the summary level Kendall's tau correlation between different metrics and human judgements.

\end{document}